\documentclass[times,twocolumn,final,authoryear,hyphens]{elsarticle}

\newcommand{\class}{c}
\newcommand{\classbinary}{y}

\newcommand{\netcliptext}{\mathcal{E}_T}
\newcommand{\netclipimage}{\mathcal{E}_I}

\newcommand{\nettransf}{\mathcal{T}}
\newcommand{\netselector}{\mathcal{S}}
\newcommand{\projection}{\mathcal{P}}

\newcommand{\methodname}{AnomalyCLIP}

\newcommand{\vectokensclass}{\boldsymbol{t}^c}
\newcommand{\vectokenscontext}{\boldsymbol{t}^\mathrm{ctx}}
\newcommand{\direction}{\boldsymbol{d}}
\newcommand{\normalmean}{\boldsymbol{m}}
\newcommand{\inputfeature}{\boldsymbol{x}}

\newcommand{\image}{\mathbf{I}}
\newcommand{\segment}{\mathbf{S}}
\newcommand{\video}{\mathbf{V}}
\newcommand{\videoanomalous}{\mathbf{V}}
\newcommand{\videonormal}{\mathbf{V}}

\newcommand{\videoanomaloustopk}{\mathcal{V}^+}
\newcommand{\videoanomalousbottomk}{\mathcal{V}^-}
\newcommand{\videonormaltopk}{\mathcal{V}^+}

\newcommand{\segmentnormaltopk}{\mathbf{S}^{+}}

\newcommand{\segmentanomaloustopk}{\mathbf{S}^{+}}
\newcommand{\segmentanomalousbottomk}{\mathbf{S}^{-}}

\newcommand{\numfeatures}{D}
\newcommand{\numframes}{F}
\newcommand{\numsegments}{S}

\newcommand{\numk}{K}
\newcommand{\numclasses}{C}
\newcommand{\batchnorm}{\mathrm{BN}}
\newcommand{\batchsize}{B}

\newcommand{\proba}{p_A}
\newcommand{\probcgivena}{p_{c|A}}
\newcommand{\probaandc}{p_{A,c}}
\newcommand{\probn}{p_N}

\newcommand{\loss}{\mathcal{L}}
\newcommand{\lossdiranomalous}{\mathcal{L}_A^\mathrm{DIR}}
\newcommand{\lossdirnormal}{\mathcal{L}_N^\mathrm{DIR}}
\newcommand{\lossdirnormaltopk}{\mathcal{L}_{N^+}^\mathrm{DIR}}
\newcommand{\lossdiranomaloustopk}{\mathcal{L}_{A^-}^\mathrm{DIR}}
\newcommand{\losstransftopk}{\mathcal{L}_{A^+}}
\newcommand{\losstransfbottomk}{\mathcal{L}_{A^-}}
\newcommand{\losstransfnorm}{\mathcal{L}_{N^+}}
\newcommand{\losstransfspars}{\mathcal{L}_\mathrm{spa}}
\newcommand{\losstransfsmooth}{\mathcal{L}_\mathrm{smo}}

\usepackage{ycviu}
\usepackage{framed,multirow}

\usepackage{amssymb}
\usepackage{latexsym}
\usepackage{amsmath}

\usepackage{url}
\usepackage{xcolor}
\definecolor{newcolor}{rgb}{.8,.349,.1}

\usepackage{xspace}
\makeatletter
\DeclareRobustCommand\onedot{\futurelet\@let@token\@onedot}
\def\@onedot{\ifx\@let@token.\else.\null\fi\xspace}

\def\eg{\emph{e.g}\onedot} 
\def\ie{\emph{i.e}\onedot}

\makeatother

\usepackage{booktabs}
\usepackage{soul}

\usepackage{array}
\usepackage{ragged2e}
\newcolumntype{P}[1]{>{\RaggedRight\hspace{0pt}}p{#1}}
\newcolumntype{X}[1]{>{\RaggedRight\hspace*{0pt}}p{#1}}
\usepackage{makecell}
\usepackage{graphicx}
\usepackage{subcaption}

\usepackage{tcolorbox}

\usepackage{tikz}
\usetikzlibrary{arrows,shapes,positioning,shadows,trees,mindmap}
\usepackage[edges]{forest}
\usetikzlibrary{arrows.meta}
\colorlet{linecol}{black!75}
\usepackage{xkcdcolors} % xkcd colors

\usepackage{setspace}

\usetikzlibrary{backgrounds}
\usetikzlibrary{arrows,shapes}
\usetikzlibrary{tikzmark}
\usetikzlibrary{calc}
\usepackage{color, colortbl}
\newcommand{\cellfirst}{\cellcolor{red!40}}
\newcommand{\cellsecond}{\cellcolor{orange!25}}
\newcommand{\cellthird}{\cellcolor{yellow!25}}

\newcommand{\eqauthor}{$^\diamond$} % Define \eqauthor as a text-based format
\newcommand{\equalcontribution}{Equal contribution.}

\usepackage{pifont}% http://ctan.org/pkg/pifont
\usepackage{tabularx}
\usepackage{enumitem}
\usepackage{color, colortbl}
\usepackage{overpic}

\usepackage[bookmarks=true,bookmarksopen=true,
           bookmarksnumbered=true,
           colorlinks=true,citecolor=black,linkcolor=black
           ]{hyperref}

\journal{Computer Vision and Image Understanding}

\begin{document}

\ifpreprint
  \setcounter{page}{1}
\else
  \setcounter{page}{1}
\fi

\begin{frontmatter}

\title{Delving into CLIP latent space for Video Anomaly Recognition}

\author[1]{Luca \snm{Zanella}\corref{cor1}\eqauthor}
\cortext[cor1]{Corresponding author:}\ead{luca.zanella-3@unitn.it}
\author[1]{Benedetta \snm{Liberatori}\eqauthor}
\author[1]{Willi \snm{Menapace}}
\author[2]{Fabio \snm{Poiesi}}
\author[2]{Yiming \snm{Wang}}
\author[1,2]{Elisa \snm{Ricci}}

\address[1]{University of Trento, Trento, Italy}
\address[2]{Fondazione Bruno Kessler, Trento, Italy}

% \received{1 May 2013}
% \finalform{10 May 2013}
% \accepted{13 May 2013}
% \availableonline{15 May 2013}
% \communicated{S. Sarkar}

\begin{abstract}
We tackle the complex problem of detecting and recognising anomalies in surveillance videos at the frame level, utilising only video-level supervision. We introduce the novel method $\methodname$, the first to combine Large Language and Vision (LLV) models, such as CLIP, with multiple instance learning for joint video anomaly detection and classification. 
Our approach specifically involves manipulating the latent CLIP feature space to identify the normal event subspace, which in turn allows us to effectively learn text-driven directions for abnormal events. 
When anomalous frames are projected onto these directions, they exhibit a large feature magnitude if they belong to a particular class. We also introduce a computationally efficient Transformer architecture to model 
short- and long-term temporal dependencies between frames, ultimately producing the final anomaly score and class prediction probabilities. We compare $\methodname$ against state-of-the-art methods considering three major anomaly detection benchmarks, \textit{i.e.}~ShanghaiTech, UCF-Crime, and XD-Violence, and empirically show that it outperforms baselines in recognising video anomalies.
Project website and code are available at \url{https://luca-zanella-dvl.github.io/AnomalyCLIP/}.
\end{abstract}

\begin{keyword}
\MSC 41A05\sep 41A10\sep 65D05\sep 65D17
\KWD Keyword1\sep Keyword2\sep Keyword3

%% MSC codes here, in the form: \MSC code \sep code
%% or \MSC[2008] code \sep code (2000 is the default)
\end{keyword}

\end{frontmatter}

\def\thefootnote{\eqauthor}\footnotetext{\equalcontribution}
%\linenumbers

%% main text
\section{Introduction}\label{sec:intro}

\begin{figure*}
    \raggedleft
    \begin{overpic}[width=.75\linewidth]{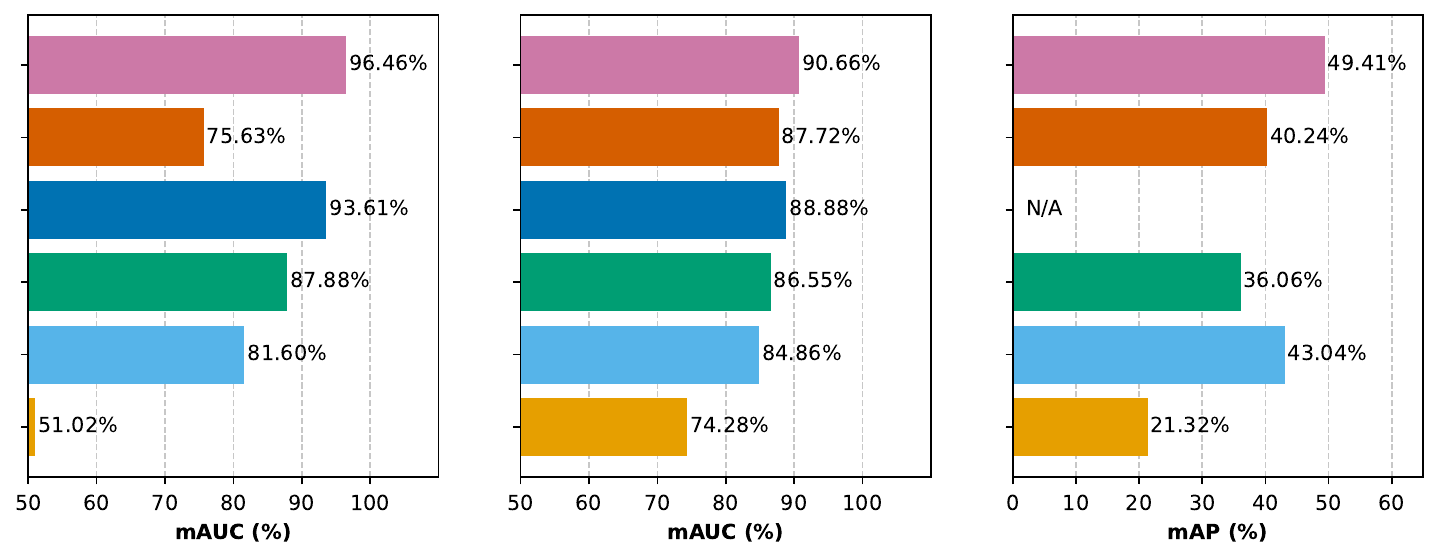}
    \put(5,40){\color{black}\scriptsize ShanghaiTech~\citep{liu2018future}}
    \put(39,40){\color{black}\scriptsize UCF-Crime~\citep{sultani2018real}}
    \put(74,40){\color{black}\scriptsize XD-Violence~\citep{wu2020not}}
    \put(-25,34){\color{black}\scriptsize AnomalyCLIP (ours)}
    \put(-25,29){\color{black}\scriptsize ActionCLIP~\citep{wang2021actionclip}}
    \put(-25,24){\color{black}\scriptsize SSRL~\citep{li2022scale}}
    \put(-25,19){\color{black}\scriptsize S3R~\citep{wu2022self}}
    \put(-25,14){\color{black}\scriptsize RTFM~\citep{tian2021weakly}}
    \put(-25,9){\color{black}\scriptsize CLIP~\citep{radford2021learning}}
    \end{overpic}
    \vspace{-3mm}
    \caption{
    Comparison of various anomaly recognition methods on the ShanghaiTech, UCF-Crime, and XD-Violence datasets in terms of the mean area under the curve (mAUC) of the receiver operating characteristic (ROC) and the mean average precision (mAP) of the precision-recall curve (PRC), which calculate the mean of binary AUC ROC and AP PRC values for all anomalous classes, respectively. A higher mAUC and mAP are crucial for video anomaly recognition as they reflect the model's ability in correctly recognising the correct abnormal class. Notably, our proposed method, \methodname, achieves the highest performance on all datasets, surpassing both the state-of-the-art methods on video anomaly detection that are re-purposed for anomaly recognition and CLIP-based video action recognition methods.}
    \label{fig:intro}
\end{figure*}
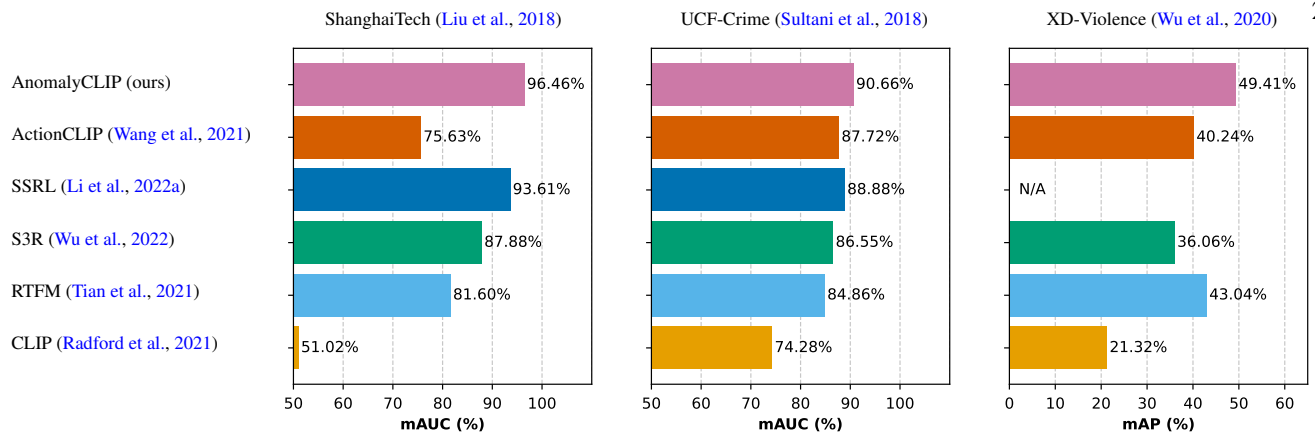

Video anomaly detection (VAD) is the task of automatically identifying activities that deviate from normal patterns in videos \citep{suarez2020survey}.
VAD has been widely studied by the computer vision and multimedia communities \citep{bao2022hierarchical,feng2021convolutional,mei2017deep, nayak2021comprehensive,sun2022evidential,wang2020cluster,xu2019video} for several important applications, such as surveillance \citep{sultani2018real} and industrial monitoring \citep{roth2022towards}.

VAD is challenging because data is typically highly imbalanced, \ie~normal events are many, whilst abnormal events are rare and sporadic.
VAD can be addressed as an out-of-distribution detection problem, \ie~one-class classification (OOC) \citep{liu2021hybrid, lv2021learning, park2020learning, xu2019video}: only visual data corresponding to the normal state is used as training data, and an input test video is classified as normal or abnormal based on its deviation from the learnt normal state.
However, OOC methods can be particularly ineffective in complex real-world applications where normal activities are diverse.
An uncommon normal activity may cause a false alarm because it differs from the learnt normal activities.
Alternatively, VAD can be addressed with fully-supervised approaches based on frame-level annotations~\citep{bai2019traffic, wang2019anomaly}.
Despite their good performance, they are considered impractical because annotations are costly to produce.
Unsupervised approaches can also be used, but their performance in complex settings is not yet satisfactory~\citep{zaheer2022generative}.
For these reasons, the most recent approaches are designed for weakly-supervised learning scenarios \citep{li2022scale, sultani2018real, tian2021weakly, wu2021learning}: they exploit video-level supervision.

Whilst existing weakly-supervised VAD methods have shown to be effective in anomaly detection~\citep{li2022scale}, they are not designed for recognising anomaly types (\eg~shooting vs.~explosion).
Performing Video Anomaly Recognition (VAR) in addition to VAD, that is not only detecting anomalous events but also recognising the underlying activities, is desirable as it provides more informative and actionable insights.
However, addressing VAR in a weakly-supervised setting is highly challenging due to the extreme data imbalance and the limited samples representing each anomaly~\citep{sultani2018real}.

We have recently experienced the emergence of powerful deep learning models that are trained on massive web-scale datasets~\citep{schuhmann2021laion}. 
These models, commonly referred to as Large Language and Vision (LLV) models or foundation models ~\citep{radford2021learning, singh2022flava}, have shown strong generalisation capabilities in several downstream tasks and have become a key ingredient of modern computer vision and multimedia systems.
These pre-trained models are publicly available and can be seamlessly integrated into any recognition system. 
LLV models can also be effectively applied to videos and to supervised action recognition tasks~\citep{wang2021actionclip,xu2021videoclip}.

In this paper, we introduce the first method that jointly addresses VAD and VAR with LLV models.
We argue that by leveraging representations derived from LLV models, we can obtain more discriminative features for recognising and classifying abnormal behaviours.
However, as supported by our experiments (Fig. \ref{fig:intro}), a naive application of existing LLV models to VAR-VAD does not suffice due to the imbalance of the training data and the subtle differences between frames of the same video containing and non containing anomalous contents.

Therefore, we propose \methodname{}, a novel solution for VAR based on the CLIP model \citep{radford2021learning}, achieving state-of-the-art anomaly recognition performance as shown in Fig.~\ref{fig:intro}. 

\methodname{} produces video representations that can be mapped to the textual description of the anomalous event.
Rather than directly operating on the CLIP feature space, we re-centre it around a normality prototype, as shown in Fig.~\ref{fig:architecture} (a). In this way, the space assumes important semantics: the magnitude of the features indicates the degree of anomaly, while the direction from the origin indicates the anomaly type. To learn the directions that represent the desired anomaly classes, we propose a Selector model that employs prompt learning and a projection operator tailored to our new space to identify the parts in a video that better match the textual description of the anomaly.
This ability is instrumental to address the data imbalance problem. We use the predictions of the Selector model to implement a semantically-guided Multiple Instance Learning (MIL) strategy that aims to widen the gap between the most anomalous segments of anomalous videos and normal ones. 
Differently from the features typically employed in VAD that are extracted using temporal-aware backbones~\citep{carreira2017quo,liu2022video}, CLIP visual features do not bear any temporal semantics as it operates at the image level.
We thus propose a Temporal model, implemented as an Axial Transformer \citep{ho2019axial}, which models both short-term relationships between successive frames and long-term dependencies between parts of the video.

As illustrated in Fig.\ref{fig:intro}, we evaluate the proposed approach on three benchmark datasets, ShanghaiTech~\citep{liu2018future}, UCF-Crime~\citep{sultani2018real} and XD-Violence~\citep{wu2020not}, and empirically show that our method achieves state-of-the-art performance in VAR.

The contributions of our paper are summarised as follows:
\begin{itemize}[noitemsep,nolistsep,leftmargin=*]
    \item we propose the first method for VAR that is based on LLV models to detect and classify the type of anomalous events;
    \item we introduce a transformation of the LLV model feature space driven by a normality prototype to effectively learn the prompt directions for anomaly types;
    \item we propose a novel Selector model that uses semantic information imbued in the transformed LLV feature space as a robust way to perform MIL segment selection and anomaly recognition;
    \item we design a Temporal model to better aggregate temporal information by modelling both the short-term relationships between neighbouring frames and the long-term dependencies among segments.
\end{itemize}

\section{Related Works}\label{sec:sota}

\noindent\textbf{Video Anomaly Detection.}
Recognising anomalous behaviours in video surveillance streams is a traditional task in computer vision and multimedia analysis. 
Existing methods for VAD can be grouped into four main categories based on the level of supervision available during training. 
The first group includes fully-supervised methods that assume available frame-level annotations in the training set \citep{bai2019traffic, wang2019anomaly}.
The second group includes weakly-supervised approaches that only require video-level normal/abnormal annotations~\citep{li2022scale, li2022self, sultani2018real, tian2021weakly, wu2021learning}. 
The third group includes one-class classification methods that assume the availability of only normal training data~\citep{liu2021hybrid, lv2021learning, park2020learning}.
The fourth group includes unsupervised models that do not use training data annotations~\citep{narasimhan2018dynamic, zaheer2022generative}.

Amongst these types of methods, weakly-supervised approaches have gained higher popularity, as they typically yield good results while limiting the annotation effort.
~\cite{sultani2018real} were the first to formulate weakly-supervised VAD as a multiple-instance learning (MIL) task, dividing each video into short segments that form a set, known as \textit{bag}.
Bags generated from abnormal videos are called positive bags, and those generated from normal videos negative bags.
Since this pioneering work, MIL has become a paradigm for VAD and several subsequent works have proposed to refine the associated ranking model to more robustly predict anomaly scores.
For example, \cite{tian2021weakly} proposed a Robust Temporal Feature Magnitude (RTFM) loss that is applied to a deep network consisting of a pyramid of dilated convolutions and a self-attention mechanism to model both short-term and long-term relationships between video snippets close in time and events in the whole video. 
\cite{wu2022self} introduced Self-Supervised Sparse Representation Learning, an approach that combines dictionary-based representation with self-supervised learning techniques to identify abnormal events.
\cite{chen2022mgfn} introduced Magnitude-Contrastive Glance-and-Focus Network, a neural network that uses a feature amplification mechanism and a magnitude contrastive loss to enhance the importance of feature discriminative for anomalies.
Motivated by the fact that anomalies can occur at any location and at any scale of the video, \cite{li2022scale} proposed Scale-Aware Spatio-Temporal Relation Learning (SSRL), an approach that extends RTFM by not only learning short-term and long-term temporal relationships but also learning multi-scale region-aware features.
While SSRL achieves state-of-the-art results in common VAD benchmarks, its high computational complexity limits its applicability.
To the best of our knowledge no previous works have explored foundation models \citep{radford2021learning} for VAD, as we propose in this work.

%%%%%%%%%%%%%%%%%%%%%%%%%%%%%%%%%%%%%%%%%%%%%%%%%%%%%%%%%%%%%%%%%%%%%%%%%%
%%%%%%%%%%%%%%%%%%%%%%%%%%%%%%%%%%%%%%%%%%%%%%%%%%%%%%%%%%%%%%%%%%%%%%%%%%
\noindent \textbf{Large Language and Vision models.}
The emergence of novel large multimodal neural networks \citep{radford2021learning, schuhmann2021laion, schuhmann2022laion, singh2022flava}, which can learn joint visual-text embedding spaces, has enabled unprecedented results in several image and video understanding tasks. 
Current LLV models adopt modality-specific encoders and are trained via contrastive techniques to align the data representations from different modalities \citep{jia2021scaling, radford2021learning}. 
Despite their simplicity, these methods have been shown to achieve impressive zero-shot generalisation capabilities.
While earlier approaches such as CLIP \citep{radford2021learning} operate on images, LLV models have recently and successfully been extended to the video domains. 
VideoCLIP~\citep{xu2021videoclip} is an example of this and it is designed to align video and textual representations by contrasting temporally overlapping
video-text pairs with mined hard negatives.
VideoCLIP can achieve strong zero-shot performance in several video understanding tasks.
ActionCLIP~\citep{wang2021actionclip} models action recognition as a video-text matching problem rather than a classical 1-out-of-N majority vote task. 
Similarly to ours, their method uses the feature space of CLIP to learn semantically-aware representations of videos. 
However, a direct exploitation of the CLIP feature space fails in capturing information on anomalous events for which a specific adaptation, proposed in this work, is necessary. In addition, action recognition methods often fall short in weakly-supervised VAD tasks due to data imbalance between normal and abnormal events, coupled with the need for frame-level evaluation at test time, despite only having video-level supervision. 
To the best of our knowledge, no prior work has specifically utilised LLV models to tackle the VAD problem.
%%%%%%%%%%%%%%%%%%%%%%%%%%%%%%%%%%%%%%%%%%%%%%%%%%%%%%%%%%%%%%%%%%%%%%%%%%
%%%%%%%%%%%%%%%%%%%%%%%%%%%%%%%%%%%%%%%%%%%%%%%%%%%%%%%%%%%%%%%%%%%%%%%%%%
%%%%%%%%%%%%%%%%%%%%%%%%%%%%%%%%%%%%%%%%%%%%%%%%%%%%%%%%%%%%%%%%%%%%%%%%%%
\section{Proposed approach}\label{sec:method}

\begin{figure*}
\centering
\begin{subfigure}{0.21\textwidth}
  \centering
  \includegraphics[width=\linewidth]{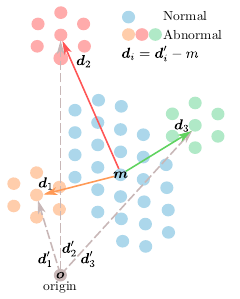}
  \caption{}
  \label{fig:sub1}
\end{subfigure}\hfill
\begin{subfigure}{0.77\textwidth}
  \centering
  \includegraphics[width=\linewidth]{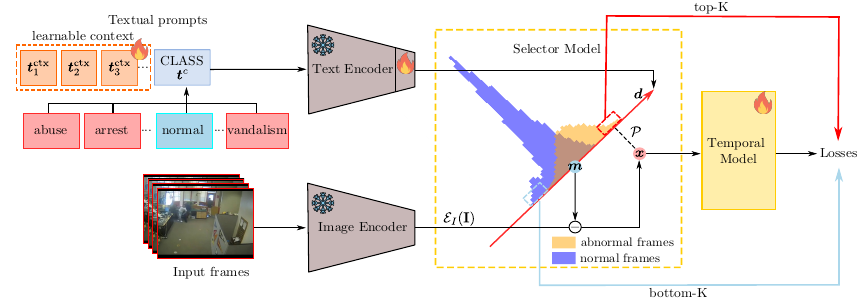}
  \caption{}
  \label{fig:sub2}
\end{subfigure}
\vspace{-2mm}
\caption{(a) Illustration of the CLIP space and the effects of the re-centring transformation with features of normal. When the space is not re-centred around the normality prototype $\normalmean$, directions $\direction'$ are similar, making it difficult to discern anomaly types, and feature magnitude is not linked to the degree of anomaly, making it difficult to identify anomalous events. When re-centred, the distribution of the magnitudes of features projected on each $\direction$ identifies the degree of detected anomaly of the corresponding type. 
(b) Illustration of our proposed framework. The Selector model learns directions $\direction$ using  CoOp \citep{zhou2022learning}, and uses them to identify the likelihood of each feature $\inputfeature$ to represent an occurrence of the corresponding anomalous class. 
MIL selection of the top-$\numk$ and bottom-$\numk$ abnormal segments is performed by considering the distribution of likelihoods along the corresponding direction. 
A Temporal model performs temporal aggregation of the features to produce the final prediction.}
\label{fig:architecture}
\end{figure*}

Weakly-supervised VAD is the task of learning to classify each frame in a video as either normal or anomalous using a dataset of tuples in the form $(\video, \classbinary)$, where $\video$ is a video and $\classbinary$ a binary label indicating whether the video contains an anomaly in any of its frames.  
With respect to VAD, in VAR we introduce the additional task of recognising the \emph{type} of the detected anomaly in each frame. 
Therefore, VAR considers a dataset of tuples $(\video, \class)$, where $\class$ indicates the type of anomaly in the video ($\class=\emptyset$ means no anomaly is present, thus being \textit{Normal}). In the following, we omit the subscripts for the purpose of readability.

To address the video-level supervision and the imbalance between normal videos and abnormal ones in VAD,
the Multiple Instance Learning (MIL) framework~\citep{sultani2018real} is widely used.
MIL models each video as a bag of segments $\video=[\segment_1,...,\segment_\numsegments] \in \mathbb{R}^{\numsegments \times \numframes \times \numfeatures}$, where $\numsegments$ is the number of segments, $\numframes$ is the number of frames in each segment, and $\numfeatures$ is the number of features associated to each frame.
Each segment can be seen as $\segment = [\inputfeature_1,...,\inputfeature_\numframes] \in \mathbb{R}^{\numframes \times \numfeatures}$ where $\inputfeature \in \mathbb{R}^{\numfeatures}$ is the feature corresponding to each frame.
MIL computes a likelihood of each frame being anomalous, selects the most anomalous ones based on it, and maximises the difference in the predicted likelihood between the normal frames and the ones selected as the most anomalous. 

In this paper, we propose to leverage the CLIP model~~\citep{radford2021learning} to address VAR and show that:

i) the alignment between the visual and textual modalities in the CLIP feature space can be used as an effective likelihood estimator for anomalies; ii) such estimator, not only can detect anomalous occurrences, but also their types; iii) such estimator is effective only when adopting our proposed CLIP space re-centring transformation (see Fig.~\ref{fig:architecture}~(a)).
Our method is composed of two models as shown in Fig.~\ref{fig:architecture}~(b): a \textit{Selector model} and a \textit{Temporal model}.
The Selector model $\netselector$ produces the likelihood that each frame belongs to an anomalous class $\netselector(\inputfeature) \in \mathbb{R}^{\numclasses}$, where $\numclasses$ is the number of anomalous classes.
We exploit the vision-text alignment in the CLIP feature space and the CoOp prompt learning approach \citep{zhou2022learning} to estimate this likelihood. 
The Temporal model $\nettransf$ assigns a binary likelihood to each frame of a video indicating whether the frame is anomalous or normal. 
Unlike $\netselector$, $\nettransf$ exploits temporal information to improve predictions and we implement it with a Transformer network~\citep{ho2019axial}. 
The predictions from $\netselector$ and $\nettransf$ are then aggregated to produce a distribution indicating the probability of a frame being normal or abnormal, and which abnormal class it belongs to.
We train our model using a combination of MIL and regularisation losses. Importantly, as $\nettransf$ is randomly initialised, the likelihood scores are less reliable, thus we always use the likelihoods produced by $\netselector$ to perform segment selection in MIL.

We describe the proposed Selector model and Temporal model in detail in Sec.~\ref{sec:selector} and Sec.~\ref{sec:temporal}, respectively. In Sec.~\ref{sec:prediction_aggregation}, we show how we aggregate the predictions of both models for estimating the final probability distribution. Finally, we describe the training and inference in Sec.~\ref{sec:training}.

%%%%%%%%%%%%%%%%%%%%%%%%%%%%%%%%%%%%%%%%%%%%%%%%%%%%%%%%%%%%%%%%%%%%%%%%%%
%%%%%%%%%%%%%%%%%%%%%%%%%%%%%%%%%%%%%%%%%%%%%%%%%%%%%%%%%%%%%%%%%%%%%%%%%%
\subsection{Selector model}
\label{sec:selector}

It is crucial for VAD and VAR to reliably distinguish anomalous and normal frames in anomalous videos given only video-level weak supervision.
Motivated by the recent findings in applying LLV models to video action recognition tasks \citep{wang2021actionclip,xu2021videoclip}, we propose a novel likelihood estimator, encapsulated by our Selector model, that combines the CLIP \citep{radford2021learning} feature space and the CoOp \citep{zhou2022learning} prompt learning approach to learn a set of directions in this space that identify each type of anomaly and their likelihood.

Our main intuition (see Fig.~\ref{fig:architecture}~(a)) is that the CLIP feature space presents an underlying structure where the set of CLIP features extracted for each frame in the dataset forms a space that is clustered around a central point which we call the normality prototype.
Consequently, the difference between a feature and the normal prototype determines important characteristics:
the magnitude of the distance reflects the likelihood of it being abnormal, while its direction indicates the type of anomaly. Such important characteristics would not be exploited by a naive application of the CLIP feature space to VAR (see Table \ref{tab:ablation-selection}). Unleashing the potential of this space in detecting anomalies thus requires a re-centring transformation, a main contribution of this work.

Following this intuition, we define the normal prototype $\normalmean$ as the average feature extracted by the CLIP image encoder $\netclipimage$ on all $N$ frames $\image$ contained in videos labelled as normal in the dataset:
%++++++++++++++++++++++++++++++++++++++++++++++++
\begin{equation}
\normalmean = \frac{1}{N} \sum_{j=1}^{N}\netclipimage(\image_j).
\end{equation}
%++++++++++++++++++++++++++++++++++++++++++++++++
For each frame $\image$ in the dataset, we produce frame features $\inputfeature$ by subtracting the normality prototype from the CLIP encoded feature, i.e., $\inputfeature=\netclipimage(\image)-\normalmean$.
 
We then exploit the visual-text aligned CLIP feature space and learn the textual prompt embeddings whose directions are used to indicate the anomalous classes.
In particular, we employ the prompt learning CoOp method~\citep{zhou2022learning} which we find ideal to find such directions as empirically demonstrated by our experiments (see Sec.~\ref{sec:exp:ablation}).

Given a class $\class$ and the textual description of the corresponding label $\vectokensclass$ expressed as a sequence of token embeddings, we consider a sequence of learnable context vectors $\vectokenscontext$ and derive the corresponding direction for the class $\direction_c \in \mathbb{R}^{\numfeatures}$ as:
%++++++++++++++++++++++++++++++++++++++++++++++++
\begin{equation}
    \direction_c=\netcliptext([\vectokenscontext,\vectokensclass]) - \normalmean,
\end{equation}
%++++++++++++++++++++++++++++++++++++++++++++++++
where $\netcliptext$ indicates the CLIP text encoder. 
The use of the textual description acts as a prior for the learned direction to match the corresponding type of anomaly, while the context vectors are jointly optimised during training as part of the parameters of $\netselector$ in order to enable the refinement of the direction. 
A different direction is learned for each class.

The learned directions serve as the base for our Selector model $\netselector$. As shown in Fig.~\ref{fig:architecture}(b), the magnitude of the projection of frame feature $\inputfeature$ on direction $\direction_c$ indicates the likelihood of the anomalous class $\class$: 
%++++++++++++++++++++++++++++++++++++++++++++++++
\begin{equation}
\netselector(\inputfeature)=[\projection(\inputfeature,\direction_1),...,\projection(\inputfeature,\direction_\numclasses)] \in \mathbb{R}^\numclasses,
\end{equation}
%++++++++++++++++++++++++++++++++++++++++++++++++
where $\projection$ indicates our projection operation. 
However, simply projecting the feature vector on the direction would make the magnitude of the projection susceptible to scale, where anomalous features of one class can potentially have a different magnitude from features of another anomalous class. To mitigate this issue, we perform a batch normalisation \citep{ioffe2015batch} after the projection which produces a distribution of projected features with zero mean and unitary variance:
%++++++++++++++++++++++++++++++++++++++++++++++++
\begin{equation}
\projection(\inputfeature,\direction_i) = \batchnorm\left(\frac{\inputfeature \cdot \direction_i}{||\direction_i ||}\right),
\end{equation}
%++++++++++++++++++++++++++++++++++++++++++++++++
where $\batchnorm$ indicates batch normalisation without affine transformation. As such, we expect within a batch the dominant normal features to be close to the origin and the abnormal features to be at the right side tail of the distribution.

The definition of likelihood can be extended to segments by summing the likelihoods of each frame:
%++++++++++++++++++++++++++++++++++++++++++++++++
\begin{equation}
\netselector(\segment)=\sum_{i=1}^{\numframes}\netselector(\inputfeature_i) \in \mathbb{R}^{\numclasses}
\end{equation}
%++++++++++++++++++++++++++++++++++++++++++++++++

%%%%%%%%%%%%%%%%%%%%%%%%%%%%%%%%%%%%%%%%%%%%%%%%%%%%%%%%%%%%%%%%%%%%%%%%%%
%%%%%%%%%%%%%%%%%%%%%%%%%%%%%%%%%%%%%%%%%%%%%%%%%%%%%%%%%%%%%%%%%%%%%%%%%%
\subsection{Temporal Model}\label{sec:temporal}

The Selector model only learns an initial \textit{time-independent} separation between anomalous and normal frames as the CLIP model operates at the image frame level. 
However, the temporal information is an important piece of information for VAR that we can exploit.
We thus propose the Temporal model $\nettransf$ to model the relationships among frames in both short-term and long-term, to enrich the visual features and to produce the predictions that indicate the likelihood of whether a frame is anomalous:
%++++++++++++++++++++++++++++++++++++++++++++++++
\begin{equation}
    \nettransf(\video) \in \mathbb{R}^{\numsegments \times \numframes}.
\end{equation}
%++++++++++++++++++++++++++++++++++++++++++++++++
We use a Transformer architecture to capture the short-term temporal dependencies between frames in a segment and the long-term temporal dependencies between all segments in a video, motivated by their success in relevant sequence modelling tasks~\citep{vaswani2017attention}.
As all the video segments of $\video$ are received as the input, the large number of segments $\numsegments$ and frames $\numframes$ increases the computational requirements for self attention. 
To reduce this cost, we implement $\nettransf$ as an Axial Transformer~\citep{ho2019axial} that computes attention separately for the two axes corresponding to the segments and the features in each segment. As suggested by experiments in Sec.~\ref{sec:exp:ablation}, Axial Transformer is also less prone to over-fitting, a likely case in VAR, as compared to standard Transformer.
We terminate the model with a sigmoid activation so that the output likelihood can also be interpreted as a probability.

%%%%%%%%%%%%%%%%%%%%%%%%%%%%%%%%%%%%%%%%%%%%%%%%%%%%%%%%%%%%%%%%%%%%%%%%%%
%%%%%%%%%%%%%%%%%%%%%%%%%%%%%%%%%%%%%%%%%%%%%%%%%%%%%%%%%%%%%%%%%%%%%%%%%%
\subsection{Predictions Aggregation}\label{sec:prediction_aggregation}

We combine the predictions from $\netselector$ and $\nettransf$ to obtain the final output: the probabilities indicating whether a frame is normal or anomalous ($\probn(\inputfeature)$ and $\proba(\inputfeature))$ and the probability that a frame presents an anomaly of a certain class ($\probaandc(\inputfeature)$).

Given an input frame feature $\inputfeature$, we define its probability of being anomalous $\proba(\inputfeature)$ as its corresponding output from the Temporal model $\nettransf$. 
The probability of the frame being normal is $\probn(\inputfeature) = 1 - \proba(\inputfeature)$. 
To obtain the probability distribution of the frame to present an anomaly of a specific class $\probaandc(\inputfeature)$, we employ the predictions of the Selector model that can be seen as the conditional distribution over the anomalous classes $\probcgivena(\inputfeature) = \mathtt{softmax}(\netselector(\inputfeature))$. 
From the definition of conditional probability it follows that $\probaandc(\inputfeature)=\proba(\inputfeature)*\probcgivena(\inputfeature)$.

%%%%%%%%%%%%%%%%%%%%%%%%%%%%%%%%%%%%%%%%%%%%%%%%%%%%%%%%%%%%%%%%%%%%%%%%%%
%%%%%%%%%%%%%%%%%%%%%%%%%%%%%%%%%%%%%%%%%%%%%%%%%%%%%%%%%%%%%%%%%%%%%%%%%%
\subsection{Training}
\label{sec:training}

%%%%%%%% MIL and LIKELIHOODS for training
We train the model following the MIL framework.
Specifically, MIL considers a batch with an equal number of normal and anomalous videos, uses the predicted likelihoods to identify the top-$\numk$ most abnormal segments in anomalous videos, and imposes separation from the other, normal ones \citep{sultani2018real}. 
Due to the higher capacity of $\nettransf$ with respect to $\netselector$ and its initial random initialisation, $\nettransf$ can not directly perform this selection since the predicted likelihoods would be excessively noisy. Instead, we use the likelihood predictions from $\netselector$ to perform MIL segment selection.

Our framework is trained end-to-end using losses on anomalous videos, losses on normal videos, and regularization losses, which we describe in the following.

%%%%%%%%%%% SELECTOR
Given an anomalous video $\videoanomalous$ of class $\class$, we define the set of top-$\numk$ most anomalous segments $\videoanomaloustopk=\{\segmentanomaloustopk_{1},...,\segmentanomaloustopk_{\numk}\}$ and, symmetrically, of bottom-$\numk$ least anomalous segments $\videoanomalousbottomk=\{\segmentanomalousbottomk_{1},...,\segmentanomalousbottomk_{\numk}\}$                      according to the likelihood assigned by the frame-level model $\netselector$ on the direction corresponding to class $\class$. We consider all frames in $\videoanomaloustopk$ and maximise the likelihood of the corresponding class being predicted by $\netselector$ by minimising the loss $\lossdiranomalous$:
%++++++++++++++++++++++++++++++++++++++++++++++++
\begin{equation}
    \lossdiranomalous=-\frac{\sum_{i=1}^{\numk} \netselector(\segmentanomaloustopk_{i})_\class}{\numk\numframes},
\end{equation}
%++++++++++++++++++++++++++++++++++++++++++++++++
where the likelihood tensor is indexed using the class $\class$.
To provide gradients to the temporal model, we also maximise $\probaandc(\inputfeature)$ for each frame contained in the segments using cross entropy:
%++++++++++++++++++++++++++++++++++++++++++++++++
\begin{equation}
    \losstransftopk=-\frac{\sum_{i=1}^{\numk}\sum_{j=1}^{\numframes} \log(\probaandc(\segmentanomaloustopk_{{i},j}))}{\numk\numframes}.
\end{equation}
%++++++++++++++++++++++++++++++++++++++++++++++++
Distinguishing normal and anomalous frames in anomalous videos is a challenging problem in VAR due to the appearance similarity between frames of the same video. To foster a better separation between these frames, we additionally consider $\videoanomalousbottomk$ and maximise $\probn(\inputfeature)$ for each frame in the segments using cross entropy:
\begin{equation}
    \losstransfbottomk=-\frac{\sum_{i=1}^{\numk}\sum_{j=1}^{\numframes} \log(\probn(\segmentanomalousbottomk_{{i},j}))}{\numk\numframes},
\end{equation}

To leverage the information in normal videos, for each segment $\segment_i$ in normal video $\videonormal$, we minimise the likelihood predicted by the Selector model:
%++++++++++++++++++++++++++++++++++++++++++++++++
\begin{equation}
    \lossdirnormal=\frac{\sum_{i=1}^\numsegments \sum_{c=1}^\numclasses \netselector(\segment_i)_\class}{\numsegments\numframes\numclasses}.
\end{equation}
%++++++++++++++++++++++++++++++++++++++++++++++++
Following the VAD literature \citep{feng2021mist, sultani2018real, tian2021weakly} we also require the model to maximise the probability of each frame in its top-$\numk$ most abnormal segments $\videonormaltopk=\{\segmentnormaltopk_{1},...,\segmentnormaltopk_{\numk}\}$ to be normal :
%++++++++++++++++++++++++++++++++++++++++++++++++
\begin{equation}
    \losstransfnorm = -\frac{\sum_{i=1}^{\numk}\sum_{j=1}^{\numframes} \log(\probn(\segmentnormaltopk_{{i},j}))}{\numk\numframes}.
\end{equation}
%++++++++++++++++++++++++++++++++++++++++++++++++

We regularise training with two additional losses \citep{sultani2018real} on all frames of anomalous videos only. 
One is a sparsity loss on the predicted scores and encourages the minimal amount of frames to be predicted as abnormal:
%++++++++++++++++++++++++++++++++++++++++++++++++
\begin{equation}
    \losstransfspars=\frac{\sum_{i=1}^\numsegments \sum_{j=1}^\numframes \proba(\video_{i,j})}{\numsegments\numframes}
\end{equation}
%++++++++++++++++++++++++++++++++++++++++++++++++
The other is a smoothness term that regularises the predictions along the temporal dimension:
%++++++++++++++++++++++++++++++++++++++++++++++++
\begin{equation}
    \losstransfsmooth=\sum_{i=2}^{\numsegments\numframes} (\proba(\video_i) - \proba(\video_{i-1})),
\end{equation}
%++++++++++++++++++++++++++++++++++++++++++++++++
where indexing is performed on the flattened sequence of frames in the video.

We jointly train the Selector and Temporal models using as final training objective: 
%++++++++++++++++++++++++++++++++++++++++++++++++
\begin{equation}
    \loss = \lossdiranomalous +  \losstransftopk + \losstransfbottomk   + \lossdirnormal + \losstransfnorm + \lambda_1\losstransfspars + \lambda_2\losstransfsmooth.
\end{equation}
%++++++++++++++++++++++++++++++++++++++++++++++++
\section{Experiments}
\label{sec:exp}

In this section, we validate our method against a range of baselines taken from state-of-the-art VAD and action recognition methods which we adapt to the VAR task. After introducing the metrics for the novel VAR task, we perform evaluation on three datasets and perform comparison in both the VAD and VAR tasks. An extensive ablation study is performed to justify our main design choices. 
Sec~\ref{sec:exp:etup} describes our experiment setup in terms of datasets and evaluation protocols. We then present and discuss the results in comparison against state-of-the-art methods in Sec~\ref{sec:exp:comparison} and the ablation study in Sec~\ref{sec:exp:ablation}.  

\subsection{Experiment Setup}\label{sec:exp:etup}
\noindent\textbf{Datasets.}
We perform our study using three widely-used VAD datasets, \ie, ShanghaiTech~\citep{liu2018future}, UCF-Crime~\citep{sultani2018real}, and XD-Violence \citep{wu2020not}.
\textit{ShanghaiTech} consists of 437 videos, recorded from multiple surveillance cameras in a university campus. A total of 130 abnormal events of 17 anomaly classes are captured in 13 different scenes. 
We adopt the dataset in the configuration of \cite{zhong2019graph} which adapts it to the weakly-supervised setting by organising it into 238 training videos and 199 testing videos.
\textit{UCF-Crime} is a large-scale dataset of real-world surveillance videos, containing 1900 long untrimmed videos that cover 13 real-world anomalies with significant impacts on public safety. The training set consists of 800 normal and 810 anomalous videos and the testing set includes the remaining 150 normal and 140 anomalous videos. 
\textit{XD-Violence} is a large-scale violence detection dataset comprising 4754 untrimmed videos with audio signals and weak labels, divided into a training set of 3954 videos and a test set of 800 videos. With a total duration of 217 hours, the dataset covers various scenarios and captures 6 categories of anomalies. Notably, each violent video may have multiple labels, ranging from 1 to 3. To accommodate our training setup, where only one anomaly type per video is considered, we select the subset of 4463 videos containing at most one anomaly.

\noindent\textbf{Performance Metrics.}
We perform evaluation in terms of both VAD and VAR. Following previous works, we measure the performance regarding VAD using the area under the curve (AUC) of the frame-level receiver operating characteristics (ROC) as it is agnostic to thresholding for the detection task. A larger frame-level AUC means a better performance in classifying between normal and anomalous events. 
To measure the VAR performance, we extend the AUC metric to the multi-classification scenario. For each anomalous class, we measure the AUC by considering the anomalous frames of the class as positive and all other frames as negatives. Successively, the mean AUC (mAUC) is computed over all the anomalous classes. Similarly, for the XD-Violence dataset, we follow the established evaluation protocol \citep{wu2020not} and present VAD results using the average precision (AP) of the precision-recall curve (PRC), while for VAR results we report the mean AP (mAP), which is calculated by averaging the binary AP values across all anomalous classes.

\noindent\textbf{Implementation details.} 
At training time, each video is divided into $\numsegments$ non-overlapping blocks. From each block, a random start-index is sampled from which segments of $\numframes$ consecutive frames are considered. If the raw video has length smaller than $\numsegments\times\numframes$, we adopt loop padding and repeat the video from the start until the minimum length of $\numsegments\times\numframes$ is reached. Each mini-batch of size $\batchsize$ used for training is composed of $\batchsize/2$ normal clips and $\batchsize/2$ anomalous clips. This is a simple but effective way to balance the mini-batch formation, which otherwise will contain mainly normal clips.
At inference, to handle videos covering arbitrary temporal windows, we first divide each video ${\video}$ into $\numsegments$ non-overlapping blocks, where each block contains frames whose number is a multiple of $\numframes$, i.e., $J\times\numframes$, where $J$ depends on the length of ${\video}$\footnote{We perform loop padding to ensure that each video is of length $J\times\numsegments\times\numframes$}. We process ${\video}$ with $J$ inferences to classify all frames in the video. At each $j^{th}$ inference, we extract the $j^{th}$ consecutive $\numframes$ frames from each block, forming segments with a total of $\numsegments\times\numframes$ that span the whole video. We then feed the segments into our approach so that our Temporal model can reason the long-term temporal relationships among segments.

For a fair comparison with previous works in VAD \citep{tian2021weakly, wu2022self, li2022scale}, we use $\numk=3$ for the MIL selection of the top-$\numk$ and bottom-$\numk$ abnormal segments, $\numsegments=32$ number of segments, $\numframes=16$ frames per segment and $\batchsize=64$ batch size. Please refer to \ref{ap:implementation_details} for more implementation details and \ref{ap:training_details} for more details on hyper-parameters.

%=======================================================
\begin{table}[t!]
\centering
\caption{Results of the state-of-the-art methods and our \methodname{} in terms of VAD and VAR on ShanghaiTech.}
\label{table:sht_results}
\tabcolsep 3pt
\vspace{-2mm}
\resizebox{\linewidth}{!}{
\begin{tabular}{llccccc}
\toprule
% \noalign{\smallskip}
Supervision & Method & Features & VAD & VAR & AUC(\%) & mAUC(\%) \\
\noalign{\smallskip}
\midrule
{} & MNAD ~\citep{park2020learning} & & \checkmark & & 70.50 & \\
\makecell[c]{One-class} & MPN ~\citep{lv2021learning} & & \checkmark & & 73.80 & \\
{} & HF$^2VAD$ ~\citep{liu2021hybrid} & & \checkmark & & 76.20 & \\
{} & \cite{zaheer2022generative} & ResNext & \checkmark & & 79.62 & \\
\midrule
{Unsupervised} & \cite{zaheer2022generative} & ResNext & \checkmark & & 78.93 \\
\midrule
\makecell[c]{Zero-shot} & CLIP  ~\citep{radford2021learning} & ViT-B/16 & & \checkmark & 49.17 & 51.02 \\
\midrule
% \noalign{\smallskip}
{} & \cite{sultani2018real} & C3D-RGB & \checkmark & & 86.30 & \\
{} & IBL ~\citep{zhang2019temporal} & C3D-RGB & \checkmark & & 82.50 & \\
{} & \cite{zaheer2022generative} & ResNext & \checkmark & & 86.21 &\\
{} & GCN ~\citep{zhong2019graph} & TSN-RGB & \checkmark & & 84.44 & \\
{} & MIST ~\citep{feng2021mist} & I3D-RGB & \checkmark & & 94.83 & \\
{} & \cite{wu2020not} & I3D-RGB & \checkmark & & & \\
Weakly- & CLAWS ~\citep{zaheer2020claws} & C3D-RGB & \checkmark & & 89.67 & \\
supervised & RTFM ~\citep{tian2021weakly} & I3D-RGB & \checkmark & & 97.21 & 81.60 \\
{} & \cite{wu2021learning} & I3D-RGB & \checkmark & & 97.48 & \\
{} & MSL ~\citep{li2022self} & I3D-RGB & \checkmark & & 96.08 & \\
{} & MSL ~\citep{li2022self} & VideoSwin-RGB & \checkmark & & 97.32 & \\
{} & S3R ~\citep{wu2022self} & I3D-RGB & \checkmark & & 97.48 & 87.88 \\
{} & MGFN ~\citep{chen2022mgfn} & I3D-RGB & \checkmark & & & \\
{} & MGFN ~\citep{chen2022mgfn} & VideoSwin-RGB & \checkmark & & & \\
{} & SSRL ~\citep{li2022scale} & I3D-RGB & \checkmark & & 97.98 & 93.61 \\
{} & ActionCLIP ~\citep{wang2021actionclip} & ViT-B/16 & & \checkmark & 96.36 & 75.63 \\
\cmidrule{2-7}
& \methodname{} (ours) & ViT-B/16 & \checkmark & \checkmark & \textbf{98.07} & \textbf{96.46} \\
\bottomrule
\end{tabular}
}
\end{table}
%=======================================================

%=======================================================
\begin{table}[t!]
\caption{Results of the state-of-the-art methods and our \methodname~in terms of VAD and VAR on UCF-Crime.}
\label{table:ucf_results}
\vspace{-3mm}
\tabcolsep 3pt
\resizebox{\linewidth}{!}{
\centering
\begin{tabular}{llccccc}
\toprule
Supervision & Method & Features & VAD & VAR & AUC(\%) & mAUC(\%) \\
\midrule
{} & SVM Baseline~\citep{sultani2018real} &  & \checkmark & & 50.00 & \\
{} & SSV ~\citep{sohrab2018subspace} &  & \checkmark & & 58.50 & \\
\makecell[c]{One-class}  & BODS ~\citep{wang2019gods} & I3D-RGB & \checkmark & & 68.26 & \\
{} & GODS ~\citep{wang2019gods} & I3D-RGB & \checkmark & & 70.46 & \\
% {} & SACR (2020)  &  & \checkmark & & 72.70 & \\
{} & \cite{zaheer2022generative} & ResNext & \checkmark & & 74.20 & \\
\midrule
\makecell[c]{Un-supervised} & \cite{zaheer2022generative} & ResNext & \checkmark & & 71.04 & \\
\midrule
\makecell[c]{Zero-shot} & CLIP  ~\citep{radford2021learning} & ViT-B/16 & & \checkmark & 58.63 & 74.28 \\
\midrule
{} & \cite{sultani2018real} & C3D-RGB & \checkmark & & 75.41 & \\
{} & \cite{sultani2018real} & I3D-RGB & \checkmark & & 77.92 & \\
{} & IBL ~\citep{zhang2019temporal} & C3D-RGB & \checkmark & & 78.66 & \\
{} & \cite{zaheer2022generative} & ResNext & \checkmark & & 79.84 & \\
{} & GCN ~\citep{zhong2019graph} & TSN-RGB & \checkmark & & 82.12 & \\
{} & MIST ~\citep{feng2021mist} & I3D-RGB & \checkmark & & 82.30 & \\
{} & \cite{wu2020not} & I3D-RGB & \checkmark & & 82.44 & \\
{} & CLAWS ~\citep{zaheer2020claws} & C3D-RGB & \checkmark & & 83.03 & \\
Weakly- & RTFM ~\citep{tian2021weakly} & VideoSwin-RGB & \checkmark & & 83.31 & \\
supervised & RTFM ~\citep{tian2021weakly} & I3D-RGB & \checkmark & & 84.03 & 84.86 \\
{} & \cite{wu2021learning} & I3D-RGB & \checkmark & & 84.89 & \\
{} & MSL ~\citep{li2022self} & I3D-RGB & \checkmark & & 85.30 & \\
{} & MSL ~\citep{li2022self} & VideoSwin-RGB & \checkmark & & 85.62 & \\
{} & S3R ~\citep{wu2022self} & I3D-RGB & \checkmark & & 85.99 & 86.55 \\
{} & MGFN ~\citep{chen2022mgfn} & VideoSwin-RGB & \checkmark & & 86.67 & \\
{} & MGFN ~\citep{chen2022mgfn} & I3D-RGB & \checkmark & & 86.98 & \\
{} & SSRL ~\citep{li2022scale} & I3D-RGB & \checkmark & & \textbf{87.43} & 88.88 \\
{} & ActionCLIP ~\citep{wang2021actionclip} & ViT-B/16 & & \checkmark & 82.30 & 87.72 \\
% {} & CLIP-TSA (2023) & CLIP & \checkmark & 87.58 & \\
\cmidrule{2-7}
 & \methodname~(ours) & ViT-B/16 & \checkmark & \checkmark & 86.36 & \textbf{90.66} \\
\bottomrule
\end{tabular}
}
\end{table}
%=======================================================

%=======================================================
\begin{table}[t!]
\caption{Results of the state-of-the-art methods and our \methodname~in terms of VAD and VAR on XD-Violence.}
\label{table:xd_violence_results}
\vspace{-3mm}
\tabcolsep 3pt
\resizebox{\linewidth}{!}{
\centering
\begin{tabular}{llccccc}
\toprule
Supervision & Method & Features & VAD & VAR & AP(\%) & mAP(\%) \\
\midrule
\makecell[c]{Zero-shot} & CLIP \citep{radford2021learning} & ViT-B/16 & & \checkmark & 27.21 & 21.32 \\
\midrule
{} & \cite{wu2020not} & C3D-RGB & \checkmark & & 67.19 & \\
{} & \cite{wu2020not} & I3D-RGB & \checkmark & & 73.20 & \\
{Weakly-} & MSL \citep{li2022self} & C3D-RGB & \checkmark & & 75.53 & \\
supervised & \cite{wu2021learning} & I3D-RGB & \checkmark & & 75.90 & \\
{} & RTFM ~\citep{tian2021weakly} & I3D-RGB & \checkmark & & 77.81 & 43.04 \\
{} & MSL \citep{li2022self} & I3D-RGB & \checkmark & & 78.28 & \\
{} & MSL \citep{li2022self} & VideoSwin-RGB & \checkmark & & 78.58 & \\
{} & S3R \citep{wu2022self} & I3D-RGB & \checkmark & & \textbf{80.26} & 36.06 \\
{} & MGFN \citep{chen2022mgfn} & I3D-RGB & \checkmark & & 79.19 & \\
{} & MGFN \citep{chen2022mgfn} & VideoSwin-RGB & \checkmark & & 80.11 & \\
{} & ActionCLIP \citep{wang2021actionclip} & ViT-B/16 & & \checkmark & 61.01 & 40.24 \\
\cmidrule{2-7}
 & \methodname~(ours) & ViT-B/16 & \checkmark & \checkmark & 78.51 & \textbf{49.41} \\
\bottomrule
\end{tabular}
}
\end{table}

%======================================================= Multiclass UCFCrime
\begin{table*}[t!]{
    \centering
    \caption{Results of the state-of-the-art methods and our \methodname{} in terms of VAR on UCF-Crime. The table highlights the top performers, with cells highlighted in \colorbox{red!40}{red} representing first place, cells in \colorbox{orange!25}{orange} representing second place, and cells in \colorbox{yellow!25}{yellow} representing third place.}
    \label{tab:ucf_multiclass_results}
    \vspace{-3mm}
    \resizebox{\textwidth}{!}{
   \begin{tabular}{lccccccccccccccc}
      \toprule
     \multirow{2}{*}{\textbf{Method}} & \multicolumn{13}{c}{\textbf{Class}} & \multirow{2}{*}{\textbf{mAUC}} \\
        \cmidrule(lr){2-14}
         & Abuse & Arrest & Arson & Assault & Burglary & Explosion & Fighting & RoadAcc. & Robbery & Shooting & Shoplifting & Stealing & Vandalism \\
        \midrule 
        RTFM~\cite{tian2021weakly} & 79.99 &  62.57 & 90.53 &  82.27 & 85.53 & 92.76 & \cellthird{85.21} & \cellthird{90.31} & 81.17 & 82.82 & \cellfirst{92.56} & 90.23 & 87.20 & 84.86\\      
        S3R~\cite{wu2022self} & \cellthird{86.38} & 68.45 & 92.19 & \cellsecond{93.55} & \cellthird{86.91} & \cellthird{93.55} & 81.69 & 85.03 & 82.07 & \cellsecond{85.32} & \cellthird{91.64} & \cellsecond{94.59} & 83.82 & 86.55 \\
        SSRL~\cite{li2022scale} & \cellfirst{95.33} & 79.26 & \cellthird{93.27} & \cellthird{91.74} & \cellsecond{89.06} & 92.25 & \cellsecond{87.36} & 80.24 & \cellfirst{87.75} & \cellthird{84.50} & \cellsecond{92.31} & \cellthird{94.22} & \cellthird{88.17} & \cellsecond{88.88} \\
        CLIP zero-shot~\cite{radford2021learning} & 57.37 & \cellthird{80.65} & \cellsecond{93.72} & 80.83 & 74.34 & 90.31 & 83.54 & 87.46 & 70.22 & 63.99 & 71.21 & 45.49 & 66.45 & 74.28 \\
        ActionCLIP~\cite{wang2021actionclip} & \cellsecond{91.88} & \cellsecond{90.47} & 89.21 & 86.87 & 81.31 & \cellsecond{94.08} & 83.23 & \cellfirst{94.34} & \cellthird{82.82} & 70.53 & 91.60 & 94.06 & \cellfirst{89.89} & \cellthird{87.72} \\ 
        \methodname & 75.03 & \cellfirst{94.56} & \cellfirst{96.66} & \cellfirst{94.80} & \cellfirst{90.08} & \cellfirst{94.79} & \cellfirst{88.76} & \cellsecond{93.30} & \cellsecond{86.85} & \cellfirst{87.45} & 89.47 & \cellfirst{97.00} & \cellsecond{89.78} & \cellfirst{90.66} \\
        \bottomrule
    \end{tabular}}}
\end{table*}

%=======================================================Multiclass Shanghaitech
\begin{table*}[t!]{
    \centering
    \caption{Results of the state-of-the-art methods and our \methodname{} in terms of VAR on ShanghaiTech. The table highlights the top performers, with cells highlighted in \colorbox{red!40}{red} representing first place, cells in \colorbox{orange!25}{orange} representing second place, and cells in \colorbox{yellow!25}{yellow} representing third place.}
    \label{tab:sht_multiclass_results}
    \vspace{-3mm}
    \resizebox{\textwidth}{!}{
    \begin{tabular}{lcccccccccccccccc}
        \toprule
        \multirow{2}{*}{\textbf{Method}} & \multicolumn{15}{c}{\textbf{Class}} & \multirow{2}{*}{\textbf{mAUC}} \\
        \cmidrule(lr){2-16}
        & Car & Chasing & Circuit & Fall & Fighting & Jumping & Monocycle  & Push & Robbery & Running & Skateboard  & Stoop & ThrowingObj. & Vaudeville & Vehicle \\
        \midrule       
        RTFM~\cite{tian2021weakly} & \cellfirst{99.70} &  95.41 & \cellthird{99.83} & 70.19 & \cellthird{97.36} & 89.14 & 37.99 & 35.28 & 67.01 &  90.59 &  96.81 & 64.11 & \cellsecond{97.93} & 91.75 & 90.85 & 81.60 \\
        S3R~\cite{wu2022self} & \cellthird{98.71} & \cellsecond{96.80} & \cellfirst{99.97} & \cellthird{85.63} & 95.93 & 69.33 & \cellsecond{96.82} & 54.76 & 61.19 & \cellthird{94.43} & 96.92 & \cellthird{75.46} & 97.63 & \cellsecond{97.78} & \cellfirst{96.84} & \cellthird{87.88} \\
        SSRL~\cite{li2022scale} & \cellsecond{99.35} & \cellfirst{97.31} & \cellsecond{99.95} & \cellsecond{91.24} & 96.88 & \cellsecond{93.07} & \cellthird{89.74} & \cellsecond{90.62} & \cellsecond{91.81} & \cellsecond{94.47} & \cellfirst{97.73} & 71.81 & \cellfirst{98.44} & 96.32 & \cellthird{95.49} & \cellsecond{93.61} \\
        CLIP zero-shot~\cite{radford2021learning} & 61.65 & 77.88 & 5.95 & 61.73 & 79.37 & 23.68 & 77.78 & \cellthird{63.36} & 37.71 & 54.39 & 76.15 & 8.47 & 44.10 & 65.97 & 27.08 & 51.02 \\
        ActionCLIP~\cite{wang2021actionclip} & 98.50	& 93.86 & 98.59	& 16.38	& \cellsecond{97.45}	& \cellthird{89.63} & \cellfirst{98.05} & 8.14 & \cellthird{67.36} & 78.25 & \cellthird{97.10} & 0.76 & \cellthird{97.70} & \cellfirst{98.65} & 93.97 & 75.63 \\
        \methodname & 98.08 & \cellthird{96.66} & 97.97 & \cellfirst{96.69} & \cellfirst{98.03} & \cellfirst{95.48} & 86.89 & \cellfirst{97.99} & \cellfirst{95.00} & \cellfirst{97.95} & \cellsecond{97.29} & \cellfirst{98.62} & 96.50 & \cellthird{96.97} & \cellsecond{96.79} & \cellfirst{96.46} \\
        \bottomrule
    \end{tabular}}}
\end{table*}
%=======================================================

%======================================================= Multiclass XDViolence
\begin{table*}[t!]{
    \caption{Results of the state-of-the-art methods and our \methodname{} in terms of VAR on XD-Violence. The table highlights the top performers, with cells highlighted in \colorbox{red!40}{red} representing first place, cells in \colorbox{orange!25}{orange} representing second place, and cells in \colorbox{yellow!25}{yellow} representing third place.}
    \label{tab:xd_multiclass_results}
    \vspace{-3mm}
    \begin{center}
    \resizebox{.6\textwidth}{!}{
    \begin{tabular}{lccccccc}
        \toprule
        \multirow{2}{*}{\textbf{Method}} & \multicolumn{6}{c}{\textbf{Class}} & \multirow{2}{*}{\textbf{mAP}} \\
        \cmidrule(lr){2-7}
         & Abuse & CarAccident & Explosion & Fighting & Riot & Shooting \\
        \midrule 
        RTFM~\cite{tian2021weakly} & \cellfirst{9.25} & \cellsecond{25.36} &	\cellthird{53.53} & \cellsecond{61.73} & \cellthird{90.38} & \cellsecond{18.01} & \cellsecond{43.04} \\
        S3R~\cite{wu2022self}  & 2.63 & 23.82 & 45.29 & 49.88 & \cellsecond{90.41} & 4.34 & 36.06 \\   
        CLIP zero-shot~\cite{radford2021learning}  & 0.32 & 12.21  & 22.26 & 25.25 & 66.60 & 1.26 & 21.32 \\ 
        ActionCLIP~\cite{wang2021actionclip}  & \cellthird{2.73} & \cellthird{25.15} & \cellsecond{55.28} & \cellthird{58.09} & 87.31	& \cellthird{12.87} & \cellthird{}40.24 \\
        \methodname & \cellsecond{6.10} & \cellfirst{31.31} & \cellfirst{68.75} & \cellfirst{71.44} & \cellfirst{92.74} & \cellfirst{26.13} & \cellfirst{49.41} \\ 
        \bottomrule 
    \end{tabular}}
    \end{center}}
\end{table*}
%=======================================================

\subsection{Evaluation Against Baselines}\label{sec:exp:comparison}

Regarding VAD, we compare \methodname~against state-of-the-art methods with different supervision setups, including one-class~ \citep{park2020learning,liu2021hybrid,lv2021learning}, unsupervised \citep{zaheer2022generative} and weakly-supervised \citep{li2022scale, tian2021weakly,wu2022self}. 
As none of the above-mentioned methods address the VAR task, we produce baselines by re-purposing some best-performing VAD methods including RTFM \citep{tian2021weakly}, S3R \citep{wu2022self} and SSRL \citep{li2022scale}\footnote{We thank authors for making their code and models publicly available}, and CLIP-based baselines~\citep{radford2021learning,wang2021actionclip}: 
\begin{itemize}[noitemsep,nolistsep]

\item \textit{Multi-classification with RTFM~\citep{tian2021weakly}, S3R~\citep{wu2022self} and SSRL~\citep{li2022scale} (weakly-supervised)}. 

We keep the original pretrained model frozen and add a multi-class classification head that we train to predict the class using a cross entropy objective on the top-$\numk$ most anomalous segments selected as in the original method. These baselines are weakly-supervised.

\item \textit{CLIP~\citep{radford2021learning} (zero-shot)}.
We achieve the classification by soft-maxing of the cosine similarities of the input frame feature $\inputfeature$ with vectors corresponding to the embedding of the textual prompt \textit{``a video from a CCTV camera of a \{class\}''} using the pre-trained CLIP model.
\item \textit{ActionCLIP~\citep{wang2021actionclip} (weakly-supervised)}. We retrain ActionCLIP~\citep{wang2021actionclip} on our datasets by propagating the video-level anomaly labels to each frame of the corresponding video.
\end{itemize}

Table~\ref{table:sht_results} presents the results on ShanghaiTech~\citep{liu2018future}. Although ShanghaiTech is a rather saturated dataset for VAD due to its simplicity in scenarios, \methodname{} scores the state-of-the-art results on both VAD and VAR, with $+0.09\%$ and $+2.85\%$ in terms of AUC ROC and mAUC ROC, respectively. ActionCLIP~\citep{wang2021actionclip} performs poorly in terms of mAUC, which we attribute to the low proportion of abnormal events in ShanghaiTech that makes the MIL selection strategy of particular importance to avoid incorrect supervisory signals on normal frames of abnormal videos.
In contrast, our proposal has a better recognition of the positive instances of abnormal videos, thus achieving better performance even when anomalies are rare. 
\methodname{} achieves a large improvement of $+45.44\%$ in terms of mAUC against zero-shot CLIP, demonstrating that a naive application of a VAR pipeline in the CLIP space does not yield satisfactory results. A revision of this space, implemented as our proposed transformation, is necessary to use it effectively.

Table~\ref{table:ucf_results} reports the results on UCF-Crime \citep{sultani2018real}. 
Our method exhibits the best discrimination of the anomalous classes, achieving the highest mAUC ROC among baselines. Similar to ShanghaiTech, it also achieves an improvement in terms of mAUC against zero-shot CLIP, verifying the importance of our proposed adaptation of the CLIP space.
Compared to ActionCLIP~\citep{wang2021actionclip}, our \methodname~obtains $+2.94\%$ in terms of mAUC, highlighting the need for a MIL framework to mitigate mis-assignment of anomalous class labels to normal frames of anomalous videos. 
It is also worth noting that the higher mAUC obtained by ActionCLIP does not result in a competitive AUC ROC on VAD, which implicates a worse separation between normal and abnormal frames.
When compared to the best performing method SSRL~\citep{li2022scale} on VAD, our method obtains an improvement of $+1.78\%$ in terms of mAUC on VAR, while being slightly worse with $-1.07\%$ in terms of AUC ROC on VAD.

Table~\ref{table:xd_violence_results} shows the results on XD-Violence \citep{wu2020not}. AnomalyCLIP outperforms other state-of-the-art methods on VAR achieving the highest mAP. Compared to the VAD baselines' models, AnomalyCLIP outperforms RTFM \citep{tian2021weakly} and demonstrates performance close to S3R \citep{wu2022self}. Please refer to \ref{ap:reproducibility_xd_violence} for further details on how we obtain results on XD-Violence.

Tables~\ref{tab:ucf_multiclass_results}, ~\ref{tab:sht_multiclass_results}, and~\ref{tab:xd_multiclass_results} display the multi-class AUC and AP for each individual abnormal class. The proposed method has a clear advantage when applied to the UCF-Crime and XD-Violence datasets, which are generally considered to be complex benchmarks in anomaly detection. Our method achieves the best mAUC and mAP on average, while it is less advantageous when dealing with anomalies that exhibit slight deviations from normal patterns, such as Shoplifting in UCF-Crime. The advantage of our proposed method is less noticeable when applied to the ShanghaiTech dataset, which captures simple scenes where most methods have achieved a saturated performance.

\begin{figure*}
\centering
\resizebox{1.0\textwidth}{!}{\includegraphics[width=\textwidth]{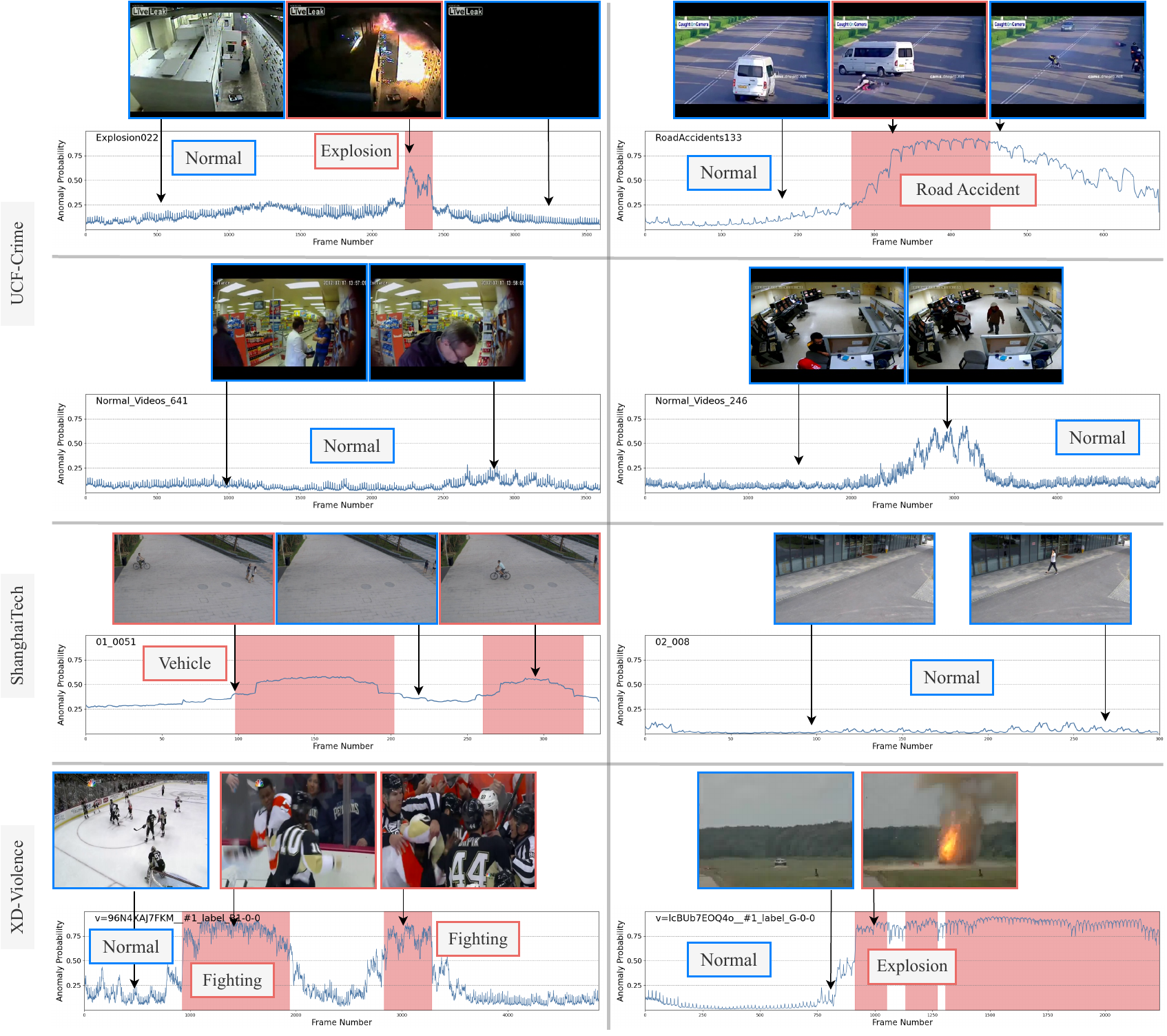}}
\caption{
Qualitative results for VAR on four test videos from UCF-Crime (the top two rows), two test videos from ShanghaiTech (the third row), and two test videos from XD-Violence (the bottom row). For each video, we show at the bottom the predicted probability of each frame being anomalous by our model over the number of frames. We showcase some key frames to reflect the relevance between the predicted anomaly probability and the visual content. The red shaded areas denote the temporal ground-truth of anomalies. We also indicate the predicted anomalous class for detected abnormal frames in the red boxes, while videos without detected anomalies are indicated with blue boxes as Normal.}
\label{fig:qualitative_var}
\end{figure*}

Fig.~\ref{fig:qualitative_var} presents the qualitative results of our proposed \methodname{} in detecting and recognising anomalies within a set of UCF-Crime, ShanghaiTech, XD-Violence test videos. The model is capable of predicting both the presence of anomalies in test videos and the category of the anomalous event. 
In video \textit{Normal\_Video\_246} from UCF-Crime (Row 2, Column 2), it can be seen how some frames have a higher-than-expected probability of being abnormal. It is interesting to note how in the video \textit{RoadAccidents133} from UCF-Crime (Row 1, Column 2) the anomaly score remains high even in the aftermath of the accident. It is also interesting to note that for Normal videos, \methodname{} is able to obtain a relatively low anomaly probability all over the frames, meaning our model has learnt a robust normal representation among Normal videos. 
Please refer to \ref{ap:qualitatives} for more results on the test videos. Furthermore, for a more intuitive understanding of the results presented in the paper, we invite readers to access the website \url{https://luca-zanella-dvl.github.io/AnomalyCLIP}, where easily accessible qualitative results are available.

\subsection{Ablation}
\label{sec:exp:ablation}

In this section, we perform ablations of our method to validate our main design choices with UCF-Crime: the way in which we represent and learn directions, the transformations applied to the CLIP space and the employed way for estimating the likelihood of anomaly, the choice of architecture for the Temporal model, training objectives, and the impact of using features extracted from different backbones.

\begin{table}[t]
\centering
\caption{Ablation on representation and learning of the directions of abnormality. `Finetuning' indicates that the last projection layer is fine-tuned. The final configuration of our model is represented by the row highlighted in \colorbox{lightgray!40}{grey} in the table.}
\label{tab:ablation-directions}
\vspace{-3mm}
{\small
\begin{tabular}{ll|cc}
\toprule
\textbf{Text encoder} & \textbf{Directions} &  \textbf{AUC} &  \textbf{mAUC}\\
\midrule
No & Direct Optimisation &  84.98 & 69.86 \\
Frozen & Engineered Prompts & 84.66 & 81.35 \\ 
Frozen & CoOp & 85.88 & 87.39 \\
\rowcolor{lightgray!40} Finetuning & CoOp &  \textbf{86.36} & \textbf{90.66} \\ 
\bottomrule
\end{tabular}
}
\end{table}

\begin{table}[t]
\centering
\caption{Comparisons of different architectural choices for
the CoOp module. `Shared' means that all the classes share a unified context, otherwise each class has a specific context. The final configuration of our model is represented by the row highlighted in \colorbox{lightgray!40}{grey} in the table.}
\label{tab:ablation-directions-coop}
\centering
\vspace{-3mm}
{\small
\centering
\begin{tabular}{cc|cc}
\toprule
\textbf{Context vectors} & \textbf{Shared}  & \textbf{AUC} &  \textbf{mAUC}\\
\midrule
4  &  &   86.16 & \textbf{91.05} \\ 
\rowcolor{lightgray!40} 8  &   & \textbf{86.36} & 90.66 \\  
16 &    & 85.82 & 90.65 \\ 
\midrule
8 & \checkmark   & 85.97 & 90.01 \\ 
\bottomrule
\end{tabular}
}
\end{table}

\noindent\textbf{Representation and Learning of the Directions.} In the ablation shown in Table \ref{tab:ablation-directions}, we evaluate the choice of the CoOp~\citep{zhou2022learning} framework to learn directions in the CLIP space. When CoOp is removed, we directly learn the directions from randomly initialized points in the CLIP space (Row 1) or make use of fixed engineered prompts of the form \textit{``a video from a CCTV camera of a \{class\}''} (Row 2). Both choices result in degradation of the results, indicating that text-guided initialization of the directions and directions finetuning are both necessary. Furthermore, we show that unfreezing the last projection of the text encoder (Row 4) enables a greater freedom in finetuning the discovered directions, yielding the best results.

In the ablation shown in Table \ref{tab:ablation-directions-coop}, we evaluate the architectural choices on the CoOp module to learn directions in the CLIP space. Specifically, we experimented by varying the number of context vectors $\vectokenscontext$ used from 4 to 8 to 16, and using shared or class-specific context vectors. Although using 4 context vectors results in a slightly higher mAUC score, we eventually opted to use 8 context vectors because they produce a higher AUC score. Results (Row 2 and 4) show that learning a specific set of context vectors for each class, is more tailored to fine-grained categories, rather than relying on more generic shared context vectors for all classes.

\begin{table}[t]
\centering
\caption{Ablation of different likelihood estimation methods, feature space transformations and MIL selection. `Features' indicates the transformation applied to CLIP features. The final configuration of our model is represented by the row highlighted in \colorbox{lightgray!40}{grey} in the table.}
\label{tab:ablation-selection}
\vspace{-3mm}
{\small
\begin{tabular}{llc|cc}
\toprule
\textbf{Likelihood} & \textbf{Features} & \textbf{MIL Selection} & \textbf{AUC} & \textbf{mAUC}\\
\midrule
cosine sim. &  CLIP  & cosine sim.  & 85.59 & 83.69 \\ 
$\netselector$ & CLIP - $\normalmean$ & feature magnitude & 84.92 & 89.82 \\ 
\rowcolor{lightgray!40} $\netselector$ & CLIP - $\normalmean$ & $\netselector$  & \textbf{86.36} & \textbf{90.66}\\
\bottomrule
\end{tabular}
}
\end{table}

\noindent\textbf{Likelihood Estimation and CLIP Latent Space Transformation.} The way in which the extracted CLIP features are transformed and the chosen likelihood estimation method play a crucial role in the quality of segment selection. We evaluate several choices in this procedure in Table \ref{tab:ablation-selection}. Directly using the CLIP space and cosine similarities with the learned directions as likelihood estimators (Row 1) produces the worst VAR results, indicating that the use of the normality prototype $\normalmean$ is of high importance in the context of anomaly detection. Second, Row 2 shows that MIL segment selection as a function of the feature magnitude without accounting for the direction is not as effective, given that the large magnitude could be attributed to irrelevant factors.

\begin{table}[t]
\centering
\caption{Comparisons of different architectural choices for the Temporal model. The final configuration of our model is represented by the row highlighted in \colorbox{lightgray!40}{grey} in the table.}
\label{tab:ablation-architecture}
\vspace{-3mm}
{\small
\begin{tabular}{l|cccc}
\toprule
\textbf{Temporal Model} & \textbf{Short-term} &  \textbf{Long-term} & \textbf{AUC} &  \textbf{mAUC}\\
\midrule
MLP  & & &  74.86 & 84.46 \\  
Transformer & \checkmark &   & 84.69 &  88.38 \\
Transformer &  & \checkmark  &  	85.10	& 89.29   \\
MTN &  &  \checkmark & 	82.71 &	87.65    \\
\rowcolor{lightgray!40} Axial Transformer & \checkmark & \checkmark  & \textbf{86.36} & \textbf{90.66}\\
\bottomrule
\end{tabular}
}
\end{table}

\begin{table}[t]
\caption{Comparisons of different architectural choices for
the Axial Transformer. The final configuration of our model is represented by the row highlighted in \colorbox{lightgray!40}{grey} in the table.}
\label{tab:ablation-axial}
\vspace{-3mm}
\centering
{\small
\centering
\begin{tabular}{ccc|cc}
\toprule
\textbf{Embedding size} & \textbf{Number of layers}  & \textbf{AUC} &  \textbf{mAUC}\\
\midrule
64 & 1  & 82.83 & 90.10 \\ 
128 & 1 & 84.97 & 90.53 \\ 
\rowcolor{lightgray!40} 256 & 1  & \textbf{86.36} & \textbf{90.66} \\ 
512 & 1  & 85.51 & 89.28 \\
\midrule
256 & 2  & 85.89 & 89.67 \\ 
256 & 3  & 85.15 & 88.14 \\ 
\bottomrule
\end{tabular}
}
\end{table}

\noindent\textbf{Temporal Model Architecture.}
Capturing temporal information is an essential aspect of VAR since it provides insights into the behaviour of objects and scenes over time.
Table \ref{tab:ablation-architecture} shows results for different architectures of $\nettransf$ \ie a 3-layer MLP, two Transformer Encoders \citep{vaswani2017attention}, the multi-scale temporal network (MTN), designed in RTFM and used in S3R and SSRL, and the employed Axial Transformer. In particular, one transformer encoder (Row 2) performs self-attention on each independent 16-frame segment, solely modelling short-term dependencies. The other (Row 3) applies self-attention on segment embeddings, which are obtained by averaging 16-frame feature embeddings within each segment, thereby only modelling long-term dependencies. To ensure a fair comparison, both transformers are designed to have a capacity similar to that of the Axial Transformer. The reduced performance of the MLP baseline (Row 1) indicates the necessity of considering temporal information which is not readily available in the extracted CLIP features. The Axial transformer can capture temporal dependencies and outperform the compared architectures. 

Table \ref{tab:ablation-axial} shows the results for different values of the embedding size and the number of layers. In the final architecture we use 1 layer and an embedding size of 256, for a total of 10.4 M trainable parameters.

\begin{table}[t]
\centering
\caption{Ablation of the losses on the Selector model. The final configuration of our model is represented by the row highlighted in \colorbox{lightgray!40}{grey} in the table.}
\label{tab:ablation-lossdir}
\vspace{-3mm}
{\small
\begin{tabular}{ccc|cc}
\toprule
\textbf{$\lossdiranomalous$} & \textbf{$\lossdirnormal$} & \textbf{AUC} & \textbf{mAUC}  \\
\midrule
& & 85.89 & 89.34 \\ 
& $\checkmark$ & 85.91 & 87.26 \\ 
$\checkmark$ & & \textbf{86.46} & \textbf{90.75} \\
\rowcolor{lightgray!40} $\checkmark$ & $\checkmark$ & 86.36 & 90.66 \\
\bottomrule
\end{tabular}
}
\end{table}

\begin{table}[t]
\centering
\caption{Ablation of losses on the aggregated outputs. The final configuration of our model is represented by the row highlighted in \colorbox{lightgray!40}{grey} in the table.}
\label{tab:ablation-losstopkbottomk}
\vspace{-3mm}
{\small
\begin{tabular}{ccc|cc}
\toprule
\textbf{$\losstransftopk$} & \textbf{$\losstransfbottomk$} & \textbf{$\losstransfnorm$} & \textbf{AUC} & \textbf{mAUC}  \\
\midrule
& $\checkmark$ & $\checkmark$ & 45.23 & 69.57 \\ 
$\checkmark$ & & $\checkmark$ & 84.50 & \textbf{90.88} \\ 
$\checkmark$ & $\checkmark$ & & 80.96 & 86.10 \\
\rowcolor{lightgray!40} $\checkmark$ & $\checkmark$ & $\checkmark$ & \textbf{86.36} & 90.66 \\
\bottomrule
\end{tabular}}
\end{table}

\begin{table}[t]
\centering
\caption{Ablation on the variation of Selector model losses. The final configuration of our model is represented by the row highlighted in \colorbox{lightgray!40}{grey} in the table.}
\label{tab:ablation-loss-S}
\vspace{-3mm}
{\small
\centering
\begin{tabular}{cccc|cc}
\toprule
\textbf{$\lossdiranomalous$} & \textbf{$\lossdirnormal$} & \textbf{$\lossdirnormaltopk$} & \textbf{$\lossdiranomaloustopk$} & \textbf{AUC} & \textbf{mAUC}  \\
\midrule
$\checkmark$ & $\checkmark$ & & $\checkmark$ & \textbf{86.41} & 88.29 \\
$\checkmark$ & & $\checkmark$ & & 86.17 & 90.53 \\
\rowcolor{lightgray!40} $\checkmark$ & $\checkmark$ & & & 86.36 & \textbf{90.66} \\
\bottomrule
\end{tabular}
}
\end{table}

\noindent\textbf{Losses.} Table \ref{tab:ablation-lossdir} illustrates the contribution of the losses on the Selector model's outputs, where we progressively remove the losses from the full training objective. The loss on abnormal videos contributes to improved VAD and VAR results on UCF-Crime. Similarly, using the loss on normal videos improves the results on Shanghaitech and XD-Violence, as can be seen in Tables \ref{tab:ablation-lossdir-sht} and \ref{tab:ablation-lossdir-xd} of \ref{ap:ablation}.

Table \ref{tab:ablation-losstopkbottomk} similarly shows the contribution of the losses on the aggregated model's output, where we remove each from the complete training objective. We validate that each of the proposed losses promotes performance on both the VAD and VAR tasks. 

The bottom-$\numk$ least anomalous segments $\videoanomalousbottomk=\{\segmentanomalousbottomk_{1},...,\segmentanomalousbottomk_{\numk}\}$ of anomalous videos proved to be beneficial for learning the Temporal Model. Inspired by this, we analyse the impact of incorporating this set of frames into the Selector Model loss by minimising the loss:
%++++++++++++++++++++++++++++++++++++++++++++++++
\begin{equation}
    \lossdiranomaloustopk=\frac{\sum_{i=1}^{\numk} \netselector(\segmentanomalousbottomk_{i})_\class}{\numk\numframes},
\end{equation}
%++++++++++++++++++++++++++++++++++++++++++++++++
Moreover, instead of using all segments of normal videos in the Selector Model loss, we evaluate the impact of using only the top-$\numk$ most abnormal segments $\videonormaltopk=\{\segmentnormaltopk_{1},...,\segmentnormaltopk_{\numk}\}$ by minimising the likelihood predicted by the Selector Model:
%++++++++++++++++++++++++++++++++++++++++++++++++
\begin{equation}
    \lossdirnormaltopk=\frac{\sum_{i=1}^{\numk} \netselector(\segmentnormaltopk_{i})_\class}{\numk\numframes}
\end{equation}
%++++++++++++++++++++++++++++++++++++++++++++++++
In Table \ref{tab:ablation-loss-S}, we present our findings, which indicate that modifying the loss function in either of two ways cause a degradation of performance. Specifically, our experiments (Row 1) demonstrate that using the bottom-$\numk$ least abnormal segments is only effective when learning the Temporal Model. This is because if there is no clear separation between the bottom-$\numk$ and top-$\numk$ abnormal features, the Selector Model can lead to incorrectly selected bottom-$\numk$ features that prevent it from learning good directions in the feature space. However, incorporating the bottom-$\numk$ least abnormal segments becomes beneficial in the Temporal Model, which has a greater capacity. Furthermore, our experiments indicate that using all normal segments (Row 3) provides a more robust estimation of the direction from normal to anomalous compared to using only the top-$\numk$ most abnormal segments (Row 2).

\begin{table}[t]
\centering
\caption{Comparisons of different features. The final configuration of our model is represented by the row highlighted in \colorbox{lightgray!40}{grey} in the table.}
\label{tab:ablation-features}
\vspace{-3mm}
{\small
\centering
\begin{tabular}{cc|cc}
\toprule
\textbf{Selector Model} & \textbf{Temporal Model} & \textbf{AUC} & \textbf{mAUC}  \\
\midrule
I3D-RGB & I3D-RGB & 65.05 & 84.24 \\
ViT-B/16 & I3D-RGB & 78.11 & 88.26 \\
ViT-B/16 & $\netselector(\inputfeature)$ & 84.44 & 86.78 \\
\rowcolor{lightgray!40} ViT-B/16 & ViT-B/16 & \textbf{86.36} & \textbf{90.66} \\
\bottomrule
\end{tabular}
}
\end{table}

\noindent\textbf{Feature Representation.} The purpose of this ablation study is to determine the most suitable feature space for the proposed method $\methodname$. To achieve this, we first investigate whether the space learned by the Selector Model can be applied to the Temporal Model. This $\numclasses$-dimensional space is formed by projecting each frame feature $\inputfeature$ onto every $\direction_c$ direction, where $\numclasses$ represents the number of anomalous classes. Our results, presented in Table \ref{tab:ablation-features}, indicate that using only this space leads to sub-optimal model performance (Row 3). This finding highlights the necessity of incorporating the information contained in the original feature space as well. We also experiment with using I3D features for both the Selector Model and the Temporal Model (Row 1), but the results demonstrate that the model using these features performs worse. We attribute this to the fact that I3D features are mapped to a region of space that is not aligned with the text features, unlike the features generated by CLIP's image encoder. For this reason, we also experimented using I3D features for the Temporal Model and features from CLIP's image encoder for the Selector Model (Row 2). The result of this experiment further emphasises that the latent space of CLIP is a more semantic space in which anomalous events of different classes are more separated, which in turn leads to superior discriminative ability in detecting and recognising anomalous events.

\section{Conclusions}
\label{sec:conclusion}
In this work, we addressed the challenging task of Video Anomaly Recognition that extends the scope of Video Anomaly Detection by further requiring 
the classification of the anomalous activities. We proposed $\methodname$, the first method that leverages LLV models in the context of VAR. Our work shed light on the fact that a naive application of existing LLV models \citep{radford2021learning,wang2021actionclip} to VAR leads to unsatisfactory performance and we demonstrated that several technical design choices are required to build a multi-modal deep network for detecting and classifying abnormal behaviours. 
We also performed an extensive experimental evaluation showing that $\methodname$
achieves state-or-the-art VAR results on the benchmark ShanghaiTech~\citep{liu2018future}, UCF-Crime~\citep{sultani2018real}, and XD-Violence~\citep{wu2020not} datasets. As future work, we plan to extend our method in open-set scenarios to reflect the real-world applications where anomalies are often not pre-defined. We will also investigate the applicability of our method in other multi-modal tasks, e.g., fine-grained classification. 

% \section*{Acknowledgments}
% Acknowledgments should be inserted at the end of the paper, before the
% references, not as a footnote to the title. Use the unnumbered
% Acknowledgements Head style for the Acknowledgments heading.

\bibliographystyle{model2-names}
\bibliography{refs}

\clearpage
% Appendix
\appendix
In this appendix, we provide further details on the implementation and training of the proposed $\methodname$. We also provide more details on how we obtain XD-Violence results. Furthermore, we report supplementary results of the ablation performed on the loss of the \emph{Selector model} to support our design choices. Lastly, we offer additional qualitative results.

\section{Implementation Details}
\label{ap:implementation_details}

Similarly to CoOp \citep{zhou2022learning}, context vectors $\vectokenscontext$ are randomly initialised by drawing from a zero-mean Gaussian distribution with standard deviation equal to $0.02$. We use the CLIP image encoder \citep{radford2021learning}, specifically the ViT-B/16 implementation, without fine-tuning, and apply standard CLIP image augmentations to each frame. As supported by the ablation in Table \ref{tab:ablation-axial}, we employ a one-layer axial transformer \citep{ho2019axial} for the Temporal Model with an embedding size of 256 for UCF-Crime~\citep{sultani2018real} and 128 for XD-Violence~\citep{wu2020not}, and a two-layer axial transformer with an embedding size of 256 for ShanghaiTech \citep{liu2018future}. In the case of UCF-Crime and XD-Violence, we use the image features of the CLIP space as input to the Temporal Model. However, for ShanghaiTech, we observe an improvement in performance by incorporating the output of the Selector Model as an additional input to the Temporal Model. This is likely because ShanghaiTech is less challenging than the other two and, as a result, the Selector Model already provides sufficient discriminative features.

Consistent with previous work on VAD \cite{tian2021weakly, wu2022self, chen2022mgfn, li2022scale}, we incorporate a random masking strategy during the selection process operated by $\netselector$. Specifically, we randomly mask 70\% of the segments to prevent the model from repeatedly selecting the same segments. This approach ensures a more diverse and representative selection of segments, thus improving the overall performance.
\section{Training Details}
\label{ap:training_details}

Training was performed using the AdamW optimiser \citep{loshchilov2017decoupled} with parameters $\beta_1=0.9$, $\beta_2=0.98$, $\epsilon=10^{-8}$ and weight decay $w = 0.2$. We tuned the learning rate and the number of epochs based on the behaviour of the training loss. Specifically, the learning rate is set to $5\times10^{-4}$, $10^{-5}$ and $5\times10^{-6}$ for ShanghaiTech, UCF-Crime, and XD-Violence, respectively, warmed up for 10\% of the total training epochs and decayed to zero following a cosine annealing schedule. The number of epochs is set to $50$ for UCF-Crime and XD-Violence, while it is set to $100$ for ShanghaiTech, due to its smaller size. We set the weight for each loss term to $1$ without tuning. Following previous work \citep{tian2021weakly, wu2022self, li2022scale}, we use $\lambda_1=8\times10^{-3}$ ad $\lambda_2=8\times10^{-4}$ for sparsity and smoothness regularisation terms, respectively.
\section{Reproducibility XD-Violence}
\label{ap:reproducibility_xd_violence}

As the original implementations of RTFM \citep{tian2021weakly} and S3R \citep{wu2022self} do not provide neither the code nor the trained models for XD-Violence \citep{wu2020not}, we made the necessary adaptations to support XD-Violence based on the information available in the original papers and on the open-source platform Github, and used the 2048-D features extracted after the final average pooling layer of the I3D ResNet50 model, pre-trained on Kinetics400 \citep{kay2017kinetics}. First, we pre-train their models on the entire XD-Violence dataset and save the checkpoint at the training iteration that obtains the highest average precision (AP) on the test set, following their training protocol. Subsequently, we maintain the original pre-trained model frozen and introduce a multiclass classification head. This newly introduced head undergoes training following the methodology outlined in Sec. \ref{sec:exp:comparison}.
\section{Ablation}
\label{ap:ablation}

\noindent\textbf{Losses.} Tables \ref{tab:ablation-lossdir-sht} and \ref{tab:ablation-lossdir-xd} illustrate the contribution of the losses on the Selector model's outputs, where we progressively remove the losses from the full training objective, on ShanghaiTech \citep{liu2018future} and XD-Violence \citep{wu2020not}, respectively. Both the losses on anomalous and normal videos contribute to better VAD and VAR results.

\begin{table}[t]
\centering
\caption{Ablation of the losses on the Selector model on ShanghaiTech \citep{liu2018future} The final configuration of our model is represented by the row highlighted in \colorbox{lightgray!40}{grey} in the table.}
\label{tab:ablation-lossdir-sht}
\vspace{-3mm}
{\small
\begin{tabular}{ccc|cc}
\toprule
\textbf{$\lossdiranomalous$} & \textbf{$\lossdirnormal$} & \textbf{AUC} & \textbf{mAUC}  \\
\midrule
& & 97.86 & 95.92 \\ 
& $\checkmark$ & 97.35 & 96.00 \\ 
$\checkmark$ & & 97.95 & 96.35 \\
\rowcolor{lightgray!40} $\checkmark$ & $\checkmark$ & \textbf{98.07} & \textbf{96.46} \\
\bottomrule
\end{tabular}
}
\end{table}

\begin{table}[t]
\centering
\caption{Ablation of the losses on the Selector model on XD-Violence \citep{wu2020not} The final configuration of our model is represented by the row highlighted in \colorbox{lightgray!40}{grey} in the table.}
\label{tab:ablation-lossdir-xd}
\vspace{-3mm}
{\small
\begin{tabular}{ccc|cc}
\toprule
\textbf{$\lossdiranomalous$} & \textbf{$\lossdirnormal$} & \textbf{AP} & \textbf{mAP}  \\
\midrule
& & 77.45 & 47.74 \\ 
& $\checkmark$ & \textbf{78.69} & 48.03 \\ 
$\checkmark$ & & 78.16 & 49.02 \\
\rowcolor{lightgray!40} $\checkmark$ & $\checkmark$ & 78.51 & \textbf{49.41} \\
\bottomrule
\end{tabular}
}
\end{table}
\section{Qualitative Results}
\label{ap:qualitatives}

\begin{figure*}
\centering
\resizebox{1.0\textwidth}{!}{\includegraphics[width=1.5\textwidth]{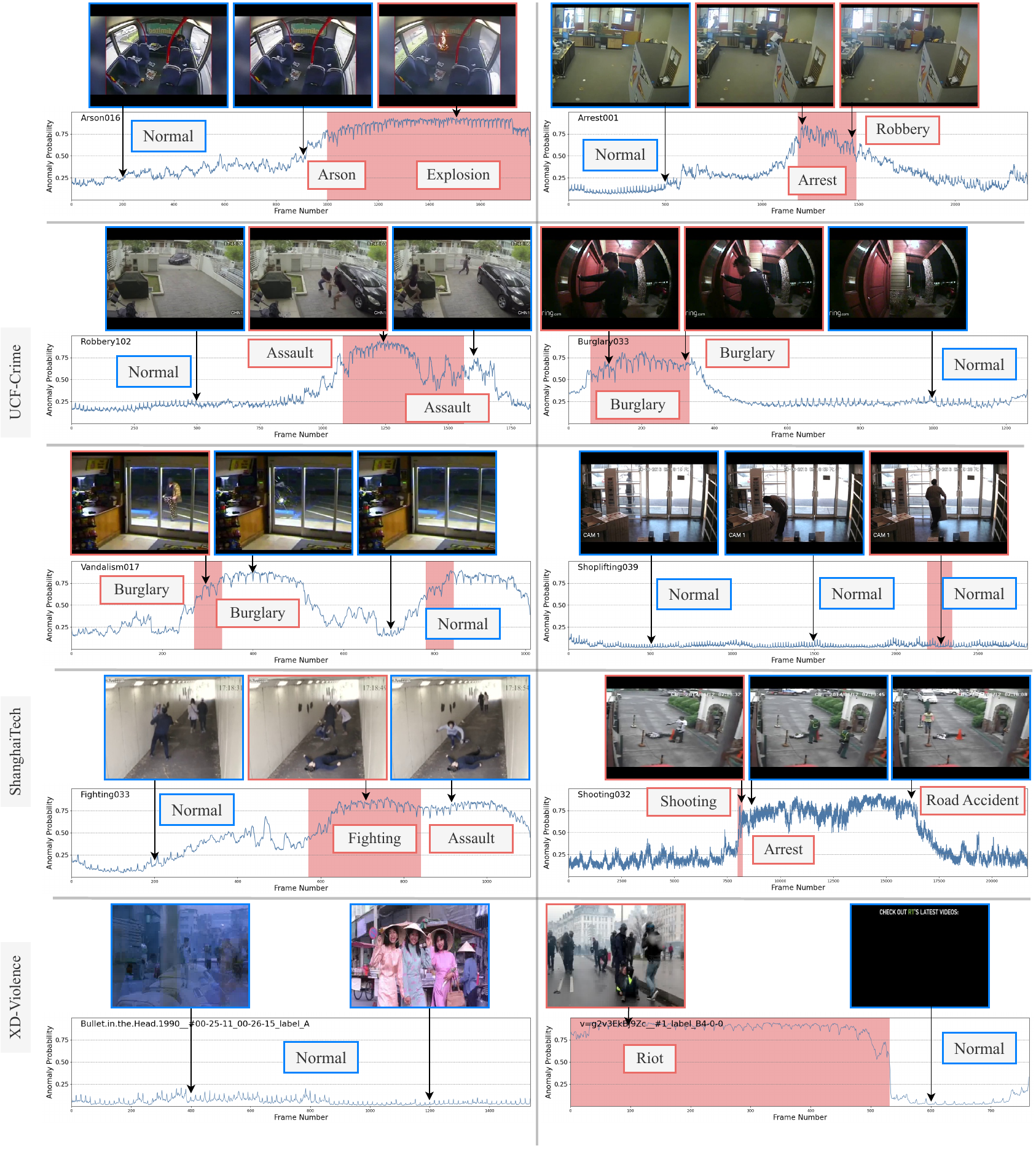}}
\caption{Qualitative results for VAR on twelve test videos from UCF-Crime (the top three rows), ShanghaiTech (the fourth row) and XD-Violence (the bottom row). For each video, we show at the bottom the predicted probability of each frame being anomalous by our model over the number of frames. We showcase some key frames to reflect the relevance between the predicted anomaly probability and the visual content. The red shaded areas denote the temporal ground-truth of anomalies. We also indicate the predicted anomalous class for detected abnormal frames in the red boxes, while videos without detected anomalies are indicated with blue boxes as Normal.}
\label{fig:qualitative_var2}
\end{figure*}

Fig. \ref{fig:qualitative_var2} presents additional qualitative results of our proposed \methodname{} in detecting and recognising anomalies within a set of UCF-Crime and ShanghaiTech test videos. The model is capable of predicting both the presence of anomalies in test videos and the category of the anomalous event. 
Videos \textit{Arson016} (Row 1, Column 1), \textit{Arrest001} (Row 1, Column 2) and \textit{Burglary033} (Row 2, Column 2) serve as good examples of the effectiveness of the proposed method. The anomalies are temporally located, and the ground-truth labels (as indicated in the video name) are correctly identified.   
However, it is worth noting that in \textit{Arson016} some frames are misjudged as Explosion, which is nevertheless a similar type of anomaly.  

One failure case is observed in the sample \textit{Shoplifting039} (Row 2, Column 2), where the proposed method fails to detect the anomaly. The reason for this failure could be attributed to the fact that the annotated anomaly is visually very similar to a normal situation, making it difficult even for humans to understand that a shoplifting is taking place and not an authorised person moving an object. This result underscores the challenge of accurately detecting anomalies in complex and visually similar scenarios. In video \textit{Robbery102} (Row 2, Column 1), the anomaly is correctly located but wrongly classified as Assault, indicating the challenges of VAR. 

Videos \textit{Shooting032} (Row 4, Column 2) and \textit{Fighting033} (Row 4, Column 1) are interesting examples that highlight the ability of the proposed method to detect anomalous situations even in the aftermath of the anomaly. In these videos, the anomaly probability remains high even after the anomalous situation annotated in the ground truth has ended, correctly indicating that there is still something anomalous happening. 

The videos from ShanghaiTech (Row 5, Columns 1-2) also provide insights into the performance of the proposed method. In the video on the left, the anomaly is correctly classified as a vehicle. However, there is also a false alarm, which represents a failure case.
On the right side of the last row, the video shows a monocycle anomaly that is wrongly classified as Running. It is reasonable to assume that the fast movement of the person riding the monocycle could have contributed to this mis-classification.

% \section*{Supplementary Material}

% Supplementary material that may be helpful in the review process should
% be prepared and provided as a separate electronic file. That file can
% then be transformed into PDF format and submitted along with the
% manuscript and graphic files to the appropriate editorial office.

\end{document}